\documentclass{article}

\usepackage[main, final]{neurips_2026}

\usepackage[utf8]{inputenc} 
\usepackage[T1]{fontenc}    
\usepackage{hyperref}       
\usepackage{url}            
\usepackage{booktabs}       
\usepackage{amsfonts}       
\usepackage{nicefrac}       
\usepackage{microtype}      
\usepackage{xcolor}         
\usepackage{amsmath}
\usepackage{multirow}
\usepackage{graphicx}
\usepackage{float}
\usepackage{wrapfig}
\usepackage{amssymb}

\newcommand{\vh}{\mathbf{h}}

\newcommand{\ve}{\mathbf{e}}

\usepackage[normalem]{ulem}

\usepackage[most]{tcolorbox}
\usepackage{inconsolata} 
\usepackage{subcaption}

\newcommand{\CoTSFT}{\textsc{CoT-SFT}}
\newcommand{\GLR}{\textsc{GLR}}
\newcommand{\GLRH}[1]{\textsc{GLR-#1}}
\newcommand{\lthead}{latent transition head}
\newcommand{\thead}{transition head}

\definecolor{tokencolor}{RGB}{0,120,130}
\definecolor{latentcolor}{RGB}{220,120,20}

\newcommand{\spTok}[1]{\textcolor{tokencolor}{\texttt{#1}}}
\newcommand{\latentsteps}[1]{%
  \begingroup
  \setlength{\fboxsep}{1pt}%
  \colorbox{orange!12}{\textcolor{latentcolor}{\texttt{\{#1 latent steps\}}}}%
  \endgroup
}

\title{Geometric Latent Reasoning Induces Shorter Generations in LLMs}

\author{
    Shashi Kumar$^{1,2,}$\thanks{Equal contribution.} \quad Yacouba Kaloga$^{1,*}$ \\ \textbf{Petr Motlicek}$^{1, 3}$ \quad \textbf{Ina Kodrasi}$^{1}$ \quad \textbf{Andrea Cavallaro}$^{2}$\\ \\ 
    $^{1}$Idiap Research Institute, Switzerland \\
    $^{2}$EPFL, Switzerland \quad
    $^{3}$BUT, Czech Republic \\
    \texttt{\{shashi.kumar, yacouba.kaloga\}@idiap.ch}
}

\begin{document}

\maketitle

\begin{abstract}
Large language models solve complex problems by generating lengthy chains of explicit reasoning tokens.
While effective, this makes reasoning expensive, length-sensitive, and constrained to (discrete) natural language.
While latent reasoning offers a continuous alternative, determining useful structures for intermediate latent states is an open challenge.
In this paper, we formulate latent reasoning as a geometric path-approximation problem within the model’s pretrained token-embedding space.
We introduce Geometric Latent Reasoning (\GLR{}), which uses a lightweight transition head to predict iterative direction updates in embedding space.
Using textual chain-of-thought traces as anchors, \GLR{} learns to approximate discrete reasoning trajectories while permitting continuous deviations from exact token embeddings.
Evaluations on mathematical reasoning benchmarks using Qwen3 models reveal an emergent phenomenon: geometric latent reasoning induces substantially shorter generations without an explicit length objective.
By replacing early explicit reasoning with continuous latent steps, models often reach correct answers using substantially fewer total generation steps.
These findings suggest that continuous trajectories act as compact intermediate reasoning states, exposing a new tradeoff between latent computation budget, output length, and accuracy.
Our code is publicly available at: \url{https://github.com/idiap/GLR}. The project page is available at: \url{https://glr-llm.github.io/}.
\end{abstract}

\section{Introduction}
\label{sec:intro}
Large language models (LLMs) increasingly rely on explicit reasoning traces, such as Chain-of-Thought (CoT), to solve complex, multi-step problems \citep{wei2022chain, yue2023mammoth, shao2024deepseekmath}. However, forcing intermediate steps into discrete natural language creates significant computational overhead. This process leads to lengthy reasoning traces and forces the model to prematurely commit to specific discrete tokens at every step. To circumvent the rigidity of text-based reasoning, latent reasoning shifts intermediate computation into continuous representation spaces. While prior approaches explore feeding unconstrained hidden states back into the model \citep{hao2024training}, distilling soft traces via auxiliary models \citep{xu2025softcot,shen2025codi}, or introducing external latent modules \citep{su2025token}, a fundamental challenge remains: determining the optimal structure for intermediate continuous states. Unconstrained states often suffer from an embedding-space mismatch, while auxiliary modules introduce complex distillation dependencies.

\begin{figure}[t]
    \centering
    \begin{minipage}{0.32\linewidth}
        \centering
        \includegraphics[width=\linewidth]{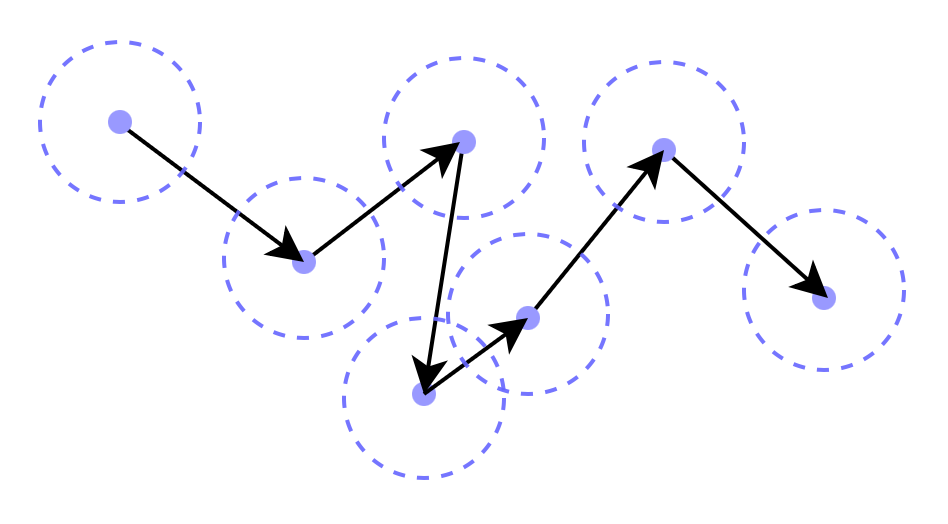}
        \vspace{-0.5em}
        \centerline{\small (a)}
    \end{minipage}
    \hfill
    \begin{minipage}{0.32\linewidth}
        \centering
        \includegraphics[width=\linewidth]{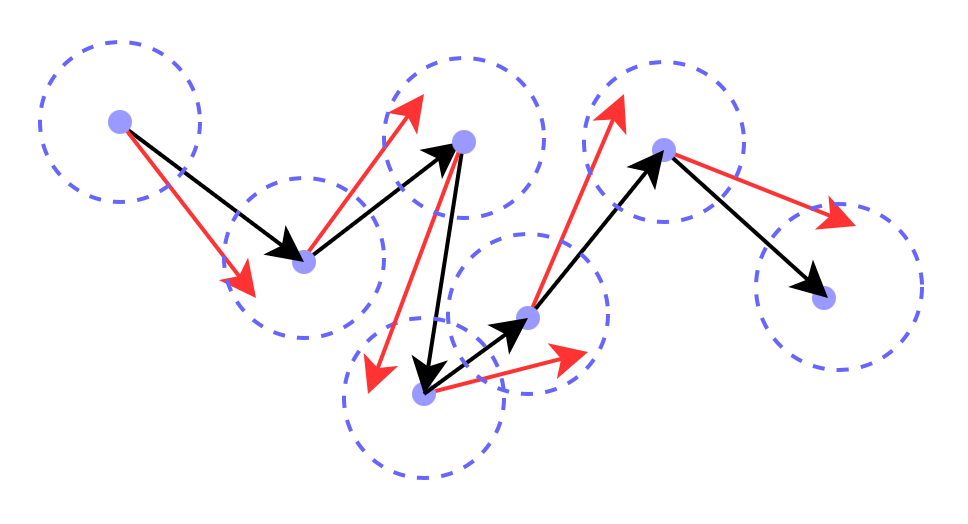}
        \vspace{-0.5em}
        \centerline{\small (b)}
    \end{minipage}
    \hfill
    \begin{minipage}{0.32\linewidth}
        \centering
        \includegraphics[width=\linewidth]{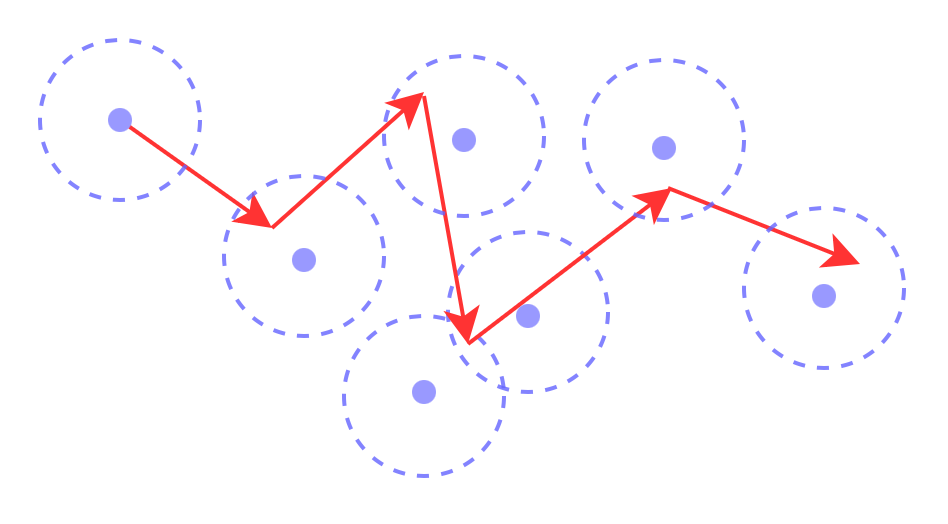}
        \vspace{-0.5em}
        \centerline{\small (c)}
    \end{minipage}

    \caption{
    \textbf{Geometric view of latent reasoning as an embedding-space trajectory.} 
\textbf{(a)} Standard chain-of-thought forces reasoning through a sequence (black arrows) of exact vocabulary embeddings (purple dots). 
    \textbf{(b)} \GLR{} learns continuous displacement vectors (red arrows) to approximate these transitions. Dashed circles denote local neighborhoods where continuous states may remain meaningful model inputs. 
    \textbf{(c)} At inference, continuous latent steps deviate from the explicit text path. By not forcing intermediate steps into discrete tokens, the model bypasses redundant transitions, taking a geometric shortcut before resuming standard token generation.
    }
    \label{fig:latent-transition}
\end{figure}

In this paper, we take a geometric view of reasoning. We first view standard textual chain-of-thought as a discrete trajectory through the model's pretrained token-embedding space (Figure~\ref{fig:latent-transition}a). This motivates our approach: rather than learning a separate unconstrained latent space, we hypothesize that useful intermediate states are not limited to exact discrete tokens, and that local neighborhoods around these trajectories can also support meaningful computation.
Our method, \textbf{Geometric Latent Reasoning (\GLR{})}, adds a lightweight \lthead{} that predicts continuous embedding-space direction updates to approximate these token-induced trajectories (Figure~\ref{fig:latent-transition}b). During training, these updates are anchored to textual CoT traces with a position-discounted mean-squared-error objective, allowing later latent states to deviate more from the original text trace.
At inference time, \GLR{} performs a fixed number of autoregressive continuous latent steps before resuming standard text token decoding (Figure~\ref{fig:latent-transition}c). Each latent state remains part of the context for subsequent computation, allowing the model to condition textual generation on the latent trajectory.
The number of latent steps controls how computation is allocated between embedding-space updates and explicit token generation, exposing a new tradeoff between latent computation budget, output length, and accuracy.

We evaluate \GLR{} on mathematical reasoning benchmarks using Qwen3 \citep{yang2025qwen3} models.
Across model sizes and benchmarks, \GLR{} induces shorter generations without an explicit length objective.
Compared with chain-of-thought supervised fine-tuning, \GLR{} often produces correct answers with substantially fewer generated tokens, especially under constrained generation budgets.
These results suggest that pretrained token-embedding spaces can support useful intermediate reasoning states beyond discrete tokens, and that geometric latent updates provide a simple mechanism for controlling the form and cost of test-time reasoning.
Our main contributions are as follows:
\begin{itemize}
\item We formulate latent reasoning as a geometric path-approximation problem within the pretrained token-embedding space, providing a structured alternative to unconstrained hidden-state feedback and unconstrained latent modules.
\item We introduce \GLR{}, a method that learns continuous embedding-space updates from textual CoT trajectories using a simple, position-discounted transition objective.
\item We show that \GLR{} induces shorter generations on mathematical reasoning benchmarks, allowing models to produce correct answers with substantially fewer generated tokens without an explicit length penalty.
\item We identify an accuracy--length tradeoff governed by the number of latent steps, exposing a new inference-time control over the allocation of computation between continuous latent updates and explicit text generation.
\end{itemize}

\section{Related Work}
\label{sec:related-works}

\paragraph{Discrete reasoning in LLMs.}
Chain-of-thought (CoT) prompting improves LLM reasoning by eliciting intermediate natural-language steps before predicting a final answer~\citep{wei2022chain}.
This behavior can be strengthened through supervised fine-tuning~\citep{yue2023mammoth}, reinforcement learning~\citep{shao2024deepseekmath}, and scaling test-time compute budgets~\citep{muennighoff2025s1,snell2024scaling}.
Notably, outcome-supervised reasoning can also increase reasoning length, as models may learn to allocate more test-time computation to improve final-answer accuracy rather than to produce shorter traces~\citep{guo2025deepseek}.
To explore alternative solutions, methods like self-consistency~\citep{wang2022self} and Tree-of-Thought~\citep{yao2023tree} sample or search over multiple textual trajectories.
However, because intermediate computation in these methods is strictly autoregressive, exploring paths requires generating multiple lengthy sequences. Furthermore, forcing every reasoning step into discrete text produces long traces that are not always faithful to the model's true internal computation~\citep{turpin2023language,lanham2023measuring}.
Our work studies a complementary direction: performing part of the intermediate computation in continuous latent states before returning to standard decoding.

\paragraph{Latent reasoning in LLMs.}
To bypass discrete token generation, recent methods explore replacing parts of explicit CoT with continuous reasoning states.
Continuous thought methods feed raw hidden states directly back into the model as subsequent inputs~\citep{hao2024training}.
Other approaches construct latent reasoning traces via knowledge distillation from teacher models~\citep{xu2025softcot,shen2025codi,deng2023implicit,xu2025softcot++} or introduce external latent modules, such as VQ-VAEs~\citep{su2025token}.
Further approaches increase latent computation through recurrent depth, repeatedly applying a shared Transformer block at inference~\citep{geiping2025scaling}, or update multiple latent thought positions in parallel using Jacobi iteration~\citep{wu-etal-2025-parallel}.
Latent reasoning was extended to visual tasks through intermediate visual representations in Machine Mental Imagery~\citep{yang2025machinementalimageryempower} and Latent Visual Reasoning~\citep{li2026latent}, and to chemical reasoning through LatentChem~\citep{ye2026latentchem}.
While these methods demonstrate the viability of non-verbalized reasoning, structuring these continuous states remains a fundamental challenge.
Unconstrained hidden states often suffer from distribution shifts when fed back as inputs, distillation pipelines introduce complex training dependencies, and external modules may not align naturally with the model's pretrained representation geometry. \GLR{} avoids these frictions by constraining latent reasoning to the model's pretrained token-embedding space.

\paragraph{Soft tokens and embedding-space explorations.}
Closest to our setting are soft-token and hybrid reasoning methods, which use continuous interpolations of token embeddings as intermediate inputs~\citep{zhang2025soft,yue2025hybrid}.
Continuous CoT also supports tracking multiple discrete reasoning traces in parallel through continuous token representations~\citep{gozeten2026continuous}.
These methods provide empirical evidence that LLMs can process continuous inputs within the embedding space without breaking, and that the token-embedding space preserves useful computational structure.
However, these approaches typically construct continuous inputs from combinations of vocabulary embeddings.
In contrast, \GLR{} treats the token-embedding space as a geometry for reasoning trajectories. Rather than taking a weighted sum of the vocabulary, it learns local directional updates to explicitly travel the embedding space.

\paragraph{Positioning.}
Our work bridges explicit CoT and latent reasoning through a geometric formulation.
Unlike text-only reasoning, which must serialize every step, and unconstrained latent methods, which operate outside the model's pretrained input geometry, \GLR{} learns continuous approximations of textual paths. The resulting method provides a lightweight way to trade explicit token generation for latent computation, yielding shorter generations without relying on explicit length penalties.

\section{Method}
\label{sec:method}
In this section, we present our latent reasoning formulation.
We first interpret textual chain-of-thought reasoning as a trajectory in the model's token-embedding space, then motivate the use of meaningful local deviations around this trajectory. We then introduce our learned latent-transition mechanism and explain how a small number of latent steps can reduce subsequent token generation.

\subsection{Preliminaries}

We use $u_{1:k}$ to denote the sequence $(u_1,\ldots,u_k)$, and $u_{<i}$
to denote the prefix $(u_1,\ldots,u_{i-1})$. Here, $k$ denotes sequence length.

\paragraph{Chain-of-thought as an embedding-space trajectory.}
Consider an input question $q_{1:n}$ for which the model generates a reasoning
trace, followed by a final answer:
\[
\texttt{<think>} \; t_{1:m} \; \texttt{</think>} \; a_{1:\ell}.
\]
Here, $t_{1:m}$ denotes the chain-of-thought tokens and $a_{1:\ell}$ denotes the answer tokens, with $n$, $m$, and $\ell$ denoting the numbers of question, reasoning, and answer tokens, respectively. At each reasoning step $i$, the model produces a hidden
state $\vh_{i}^t \in \mathbb{R}^d$ from the current context $(q_{1:n}, t_{<i})$.
This hidden state induces a distribution over the vocabulary,
\[
p_\theta(\cdot \mid q_{1:n}, t_{<i})
=
\mathrm{softmax}(W_{\mathrm{out}} \vh_{i}^t),
\]
where $W_{\mathrm{out}}$ is the vocabulary projection matrix. The next thought token $t_i$ is sampled as:
\[
t_i \sim p_\theta(\cdot \mid q_{1:n}, t_{<i}).
\]
Once selected, $t_i$ is mapped to its input embedding
\[
\ve_i^t = E_{\mathrm{in}}(t_i),
\]
which is fed back into the model to produce the next hidden state
$\vh_{i}^t$. Thus, although the visible chain-of-thought is a discrete
sequence of tokens, it induces a sequence of input embeddings
\[
\ve_1^t, \ve_2^t, \ldots, \ve_m^t.
\]
We view this sequence as a reasoning trajectory in the model's embedding
space (Figure~\ref{fig:latent-transition}a). Under this perspective, searching for better reasoning can also be
interpreted as searching over trajectories in the continuous input space
through which the model performs reasoning.

\paragraph{Local continuity of embedding-space reasoning states.}
This trajectory view is useful only if neighborhoods around token embeddings
can remain meaningful inputs to the model (Figure~\ref{fig:latent-transition}b). A minimal justification comes from
the model itself: for a fixed prefix $(q_{1:n}, t_{<i})$, the Transformer
defines a continuous map
$F_\theta(\cdot \mid q_{1:n}, t_{<i}) : \ve_i^t \mapsto \vh_{i}^t$
from the current input embedding to the corresponding hidden state. Therefore,
a small perturbation $\delta \ve$ of an embedding is expected to produce a nearby hidden
state,
\[
F_\theta(\ve_i^t + \delta \ve \mid q_{1:n}, t_{<i})
\approx
F_\theta(\ve_i^t \mid q_{1:n}, t_{<i})
\quad \text{for small } \|\delta \ve\|_2.
\]
Since this hidden state is then projected to the vocabulary distribution,
nearby hidden states are expected to induce similar predictive behavior.

This continuity argument does not imply that arbitrary embedding-space points
are useful. However, soft-token and continuous-thinking methods \citep{zhang2025soft,xu2025softcot} provide direct
empirical evidence that language models can process continuous inputs that do
not correspond to a single discrete token. Instead of feeding back a sampled
token embedding $\ve^t_i = E_{\mathrm{in}}(t_i)$, these methods may feed a soft
embedding
\[
\tilde{\ve}^t_i
=
\sum_{v \in \mathcal{V}}
p_\theta(v \mid q_{1:n}, t_{<i}) E_{\mathrm{in}}(v),
\qquad
\sum_{v \in \mathcal{V}} p_\theta(v \mid q_{1:n}, t_{<i}) = 1.
\]
Although $\tilde{\ve}_i^t$ generally does not equal any single vocabulary
embedding, it can still support coherent reasoning when fed back into the
model. This supports the hypothesis that neighborhoods around token-induced
embedding trajectories contain meaningful intermediate latent states.

\subsection{Geometric Latent Reasoning}
\label{sec:glr}
Rather than searching over arbitrary latent states, we learn local transitions around token-induced embedding trajectories.
The goal is to keep latent reasoning within the pretrained input geometry while allowing the learned transition to move toward continuous states that improve subsequent token prediction.

\begin{figure}[t]
    \centering
    \includegraphics[width=0.9\linewidth]{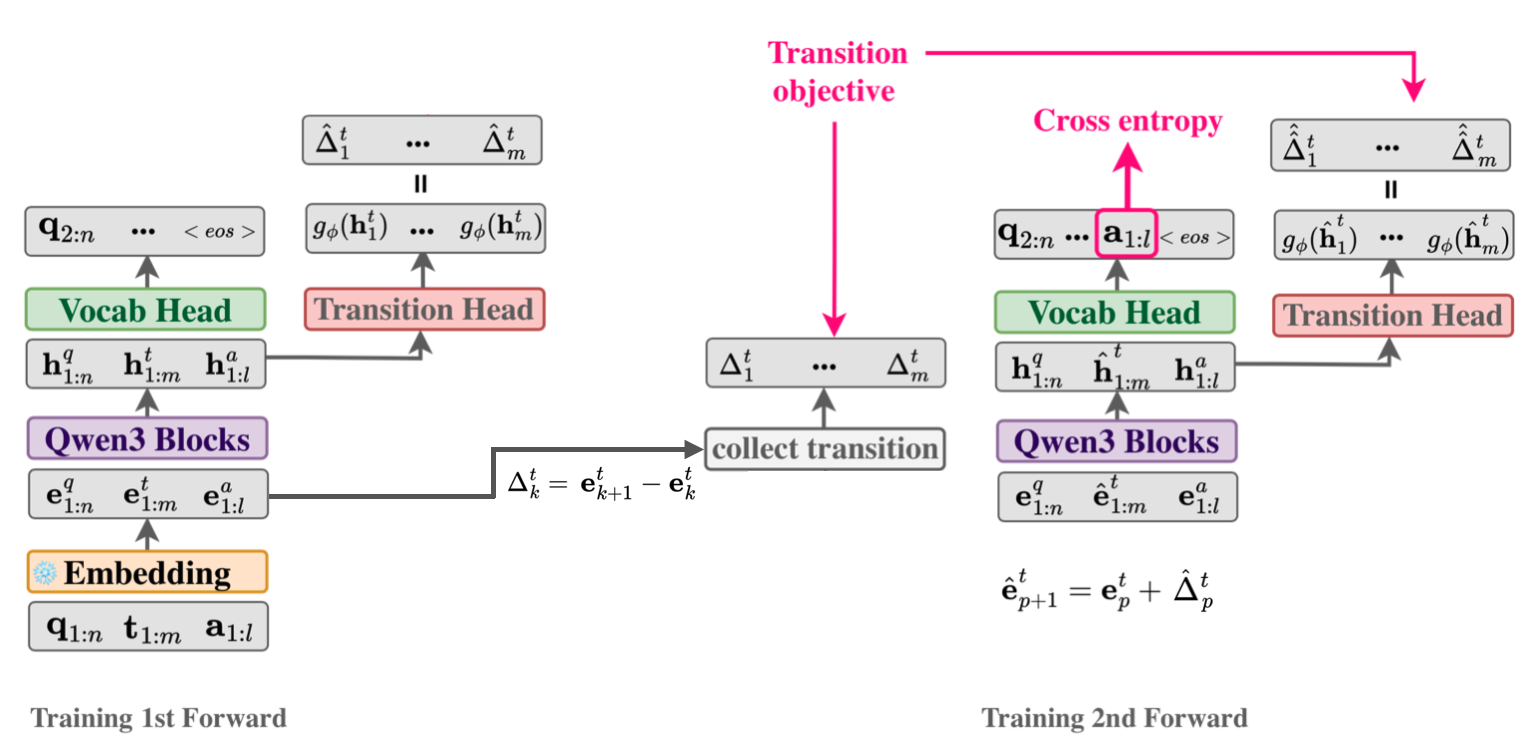}
   \caption{\textbf{Training pipeline for Geometric Latent Reasoning (\GLR{}).} \textbf{Left (first forward pass):} The model processes the original discrete sequence to collect both the exact ($\Delta_k^t$) and predicted ($\hat{\Delta}_k^t$) embedding-space displacements between consecutive reasoning tokens. \textbf{Right (second forward pass):} Discrete thought embeddings ($\mathbf{e}^t$) are replaced with continuous latent states ($\hat{\mathbf{e}}^t$) obtained by applying the Transition Head output from the first pass (i.e., $\hat{\Delta}_k^t$). The model then processes this modified sequence to compute the final objectives: standard cross-entropy on the answer tokens ($\mathbf{a}_{1:l}$), preserving generation capabilities, and a transition objective that anchors the continuous latent updates (i.e., the second Transition Head output, $\hat{\hat{\Delta}}_k^t$) to the true discrete trajectory update ($\Delta_k^t$).}
    \label{fig:main-figure}
\end{figure}
\paragraph{Learning latent transitions.}
We add to the language model a lightweight \lthead{} (Figure~\ref{fig:main-figure})
\[
g_\phi : \mathbb{R}^d \rightarrow \mathbb{R}^d,
\]
implemented as a linear layer on top of the model hidden states.
Although the latent trajectory is defined in the input token-embedding space, the transition is predicted from the final hidden state associated with the current reasoning context. Given
$\vh_{i-1}^t$, the head predicts an embedding-space update
\[
\Delta \hat\ve_i^t = g_\phi(\vh_{i-1}^t).
\]
Rather than predicting an arbitrary next embedding, the head predicts a local displacement along the token-induced embedding trajectory. Let
\[
\ve_i^t = E_{\mathrm{in}}(t_i)
\]
be the embedding of the $i$-th thought token. We define the target transition as the difference between consecutive reasoning-token embeddings:
\[
\Delta \ve_i^t = \ve_i^t - \ve_{i-1}^t,
\]
where $\ve_0$ denotes the embedding immediately preceding the first thought
token, e.g. the embedding of \texttt{<think>}. The \thead{} is trained
to approximate this displacement using a position-discounted Mean Squared Error:
\[
\mathcal{L}_{\Delta}
=
\frac{1}{m}
\sum_{i=1}^{m}
\gamma^{i-1}
\left\|
g_\phi(\vh_{i-1}^t) - \Delta \ve_i^t
\right\|_2^2,
\]
where \(0 < \gamma \leq 1\) is a discount factor that reduces the transition penalty for later reasoning positions.

\paragraph{Training with latent replacements.}
To ensure the model can robustly condition on continuous inputs without representation collapse, we train \GLR{} using a two-pass procedure (Figure~\ref{fig:main-figure}).
Let
$s_{1:N}=(q_{1:n},\texttt{<think>},t_{1:m},\texttt{</think>},a_{1:\ell})$
denote the full supervised sequence.
In the first forward pass, the model processes the original discrete
sequence and predicts an embedding-space transition for each discrete reasoning
position. For each thought token $t_i$, we construct a latent replacement
\[
\hat\ve_i
=
\ve_{i-1} + g_\phi(\vh_{i-1}^t).
\]
The embeddings inside the \texttt{<think>} span are replaced by these latent
embeddings, and the model is run again on the modified sequence. This second
forward pass is used to compute the latent transition objective; no
cross-entropy (CE) loss is applied to the replaced reasoning tokens. The final
objective is
\[
\mathcal{L}
=
\mathcal{L}_{\mathrm{CE}}
+
\lambda \mathcal{L}_{\Delta},
\]
where $\mathcal{L}_{\mathrm{CE}}$ is applied only to the textual answer tokens $a_{1:\ell}$, while $\mathcal{L}_{\Delta}$ trains the transition head to follow local movements along the token-induced embedding trajectory.
This decoupling reflects the
role of the two objectives: the CE loss maintains the standard
language-modeling behavior, while the transition loss learns how to move within
the continuous embedding space already shaped by pretraining.

Note that we do not apply token-level CE to the latent replacement positions. 
The latent states in \GLR{} are not intended to be independently verbalizable tokens; they are continuous intermediate states whose utility is measured by their effect on downstream answer generation. 
In preliminary experiments, applying CE to latent positions degraded performance, likely because it forces each latent state back toward immediate vocabulary prediction and counteracts the geometric relaxation introduced by the transition objective.

\paragraph{Latent reasoning at inference.}
At inference time, once the model enters the reasoning span, we choose latent
steps $K$ that specify how many reasoning steps are performed in continuous space before returning to standard token decoding. Let
$\widetilde{\ve}_1 = E_{\mathrm{in}}(\texttt{<think>})$ initialize the latent
reasoning trajectory. For $i=1,\ldots,K$, the model predicts a transition from
the current hidden state $\vh_i$ and updates the latent input as
\[
\hat\ve_{i} =  \hat\ve_{i-1} + g_\phi(\vh_{i-1}^t).
\]
The resulting continuous embedding is fed directly back into the model instead of the embedding of a sampled token. Each latent state is generated autoregressively, conditioned on the prompt and all preceding latent states, and remains part of the context for subsequent latent and textual decoding. After $K$ latent steps, the model resumes token-level reasoning and answer generation conditioned on the accumulated context.

Thus, \GLR{} moves through the embedding space without forcing every
intermediate step to correspond to a vocabulary token. This may allow the model to bypass transitions that are useful for (readable) chain-of-thought but
may carry little reasoning content, thereby reducing the amount of explicit
reasoning text needed before answer generation.

\section{Experiments}
\label{sec:experiments}
Our experiments investigate how Geometric Latent Reasoning (\GLR{}) alters the allocation of computation between continuous latent transitions and explicit token generation. We aim to answer three questions: (1) Does \GLR{} improve accuracy under strictly constrained generation budgets? (2) Does the geometric objective reduce the total number of sequential steps required to solve a problem? (3) How does the latent-step budget $K$ dictate the tradeoff between generation length and accuracy?

\subsection{Setup}
\label{sec:exp-setup}

\paragraph{Models and training data.}
We evaluate \GLR{} using Qwen3-0.6B and Qwen3-1.7B \citep{yang2025qwen3}. Both models are initialized from their pretrained checkpoints to ensure a well-formed token-embedding space. For \GLR{}, the \lthead{} ($g_\phi$) is initialized from scratch, adding approximately 1M and 4M trainable parameters to the 0.6B and 1.7B models, respectively. These models are fine-tuned on a randomly sampled 10K-example subset of the \texttt{math} split from the Open-R1 Mixture-of-Thoughts dataset \citep{openr1,lozhkov2025openr1math220k}, which provides high-quality textual chain-of-thought traces. We filter out examples exceeding 8,192 tokens.  Additional details in Appendix~\ref{app:datasets-train}.

\paragraph{Training configurations.}
We compare \GLR{} against a standard supervised fine-tuning baseline (\CoTSFT{}), trained using the standard next-token cross-entropy objective over the same 10K CoT traces. For \GLR{}, the model is augmented with the \lthead{} and trained using the two-pass procedure described in Section~\ref{sec:method}. Crucially, we freeze the input token-embedding layer for all experiments. Allowing embeddings to update while $g_\phi$ simultaneously learns to predict displacements between them creates a non-stationary target that destabilizes training \citep{mnih2015human,he2020momentum}. To ensure a fair comparison, the embedding layer is also frozen for the \CoTSFT{} baseline. Additional training hyperparameters are detailed in Appendix~\ref{app:training-details}.

\paragraph{Evaluation setup.}
We evaluate pass@1 accuracy under greedy decoding across six mathematical benchmarks.
We primarily analyze the accuracy--length frontier on GSM8K~\citep{cobbe2021training} for foundational arithmetic and MATH500 \citep{hendrycks2021measuring} for complex derivations. To assess generalization, we evaluate on MultiArith~\citep{roy2015solving}, AMC23~\citep{yang2024qwen2}, and OlympiadBench~\citep{he2024olympiadbench}. Crucially, we include SVAMP~\citep{patel2021nlp}, a set of highly simplified arithmetic problems, to observe whether \GLR{} bypasses redundant reasoning traces. Accuracy is computed using the \texttt{lm-evaluation-harness} framework \citep{eval-harness}. Additional details are provided in Appendix~\ref{app:datasets-eval}.
At inference, \GLRH{$K$} executes $K$ continuous latent steps before resuming standard autoregressive token decoding. For Qwen3-0.6B, we evaluate $K \in \{5, 10, 20, 50\}$. For Qwen3-1.7B, we expand this to $K \in \{5, 10, 20, 50, 80, 100\}$, hypothesizing that larger models possess more expressive embedding spaces capable of supporting longer continuous trajectories.

Decoding is constrained to fixed maximum generation limits: 2048 steps for arithmetic benchmarks (GSM8K, SVAMP, MultiArith) and 4096 steps for advanced reasoning (MATH500, AMC23, OlympiadBench). To measure efficiency, we define generation length as all model steps after the prompt.
For \CoTSFT{}, this counts all generated text tokens. For \GLR{}, this counts the $K$ latent steps plus all subsequent text tokens. This is a conservative accounting for \GLR{}: each latent step still requires a Transformer forward pass, but bypasses the vocabulary projection, instead applies the \thead{}.
For Qwen3-1.7B, latent rollout replaces a roughly 300M-weight vocabulary projection with a 4M-parameter \thead{}.
We additionally measure wall-clock latency for both methods using the same HuggingFace~\citep{wolf2019huggingface} decoder; Appendix~\ref{app:compute-accounting} provides the measurement protocol and full efficiency results.
Finally, when plotting length distributions, we measure only successful generations to isolate the minimal active steps required to reach a correct answer.

\subsection{Results: Accuracy and Generation Length}
\label{sec:res-disc}
\begin{figure}[t]
    \centering
    \includegraphics[width=0.95\linewidth]{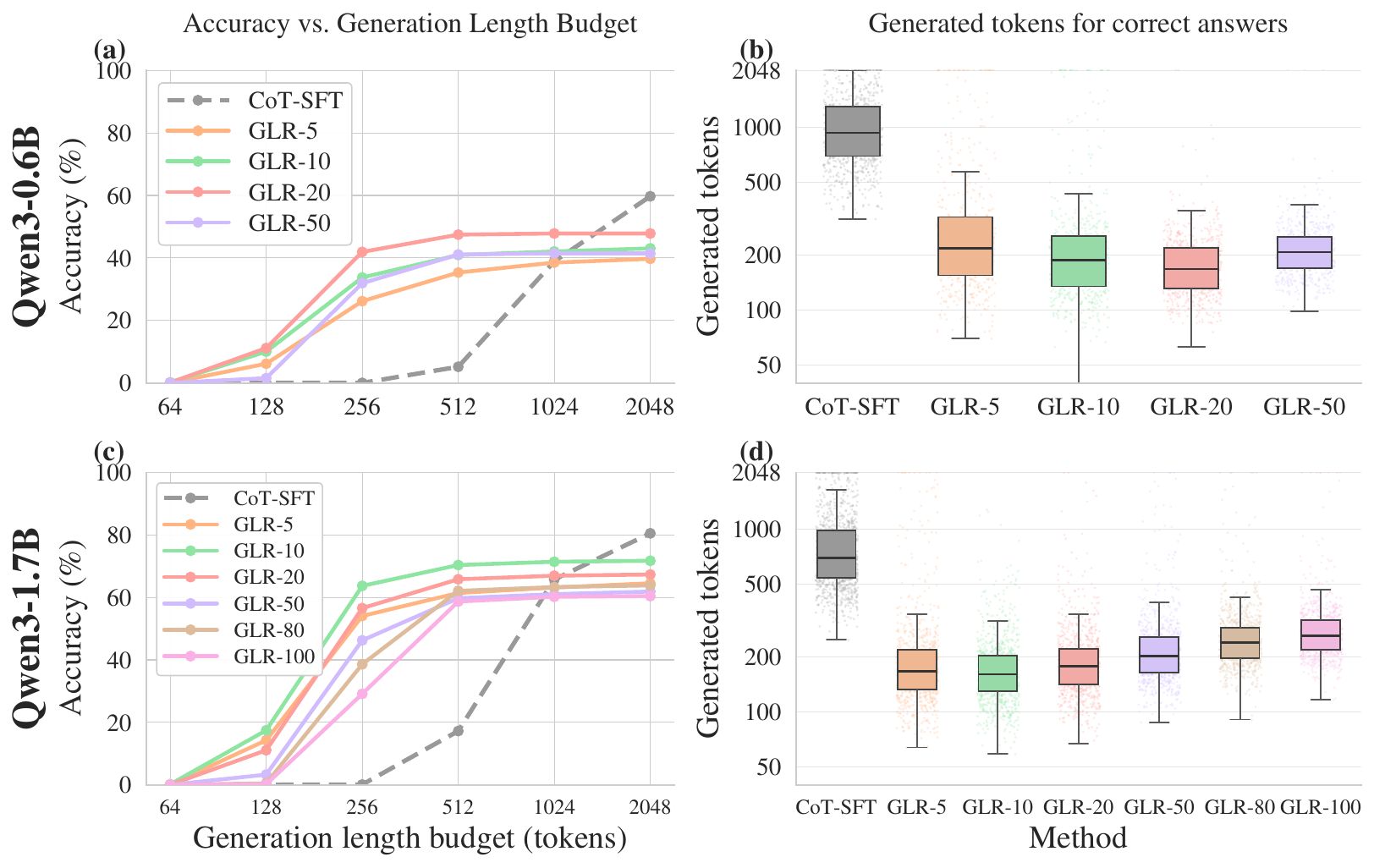}
    \caption{
    \textbf{\GLR{} shifts the accuracy--length frontier and reduces generation length on GSM8K.} 
    Models are fine-tuned on same training sets.
    \textbf{Left:} Pass@1 accuracy as a function of the generation length budget.  \GLR{} improves accuracy under constrained budgets compared to standard CoT (\CoTSFT{}).
    \textbf{Right:} Distribution of total generated steps for correct answers. \GLR{} reaches the correct solution in substantially fewer total steps. 
    \GLRH{$K$} denotes inference with $K$ latent steps. Right y-axes are log-scaled; points at 2048 indicate truncated generations.
    }
    \label{fig:gsm8k-qwen3-combined}
    \vspace{-0.5cm}
\end{figure}

\begin{figure}[t]
    \centering
    \includegraphics[width=0.95\linewidth]{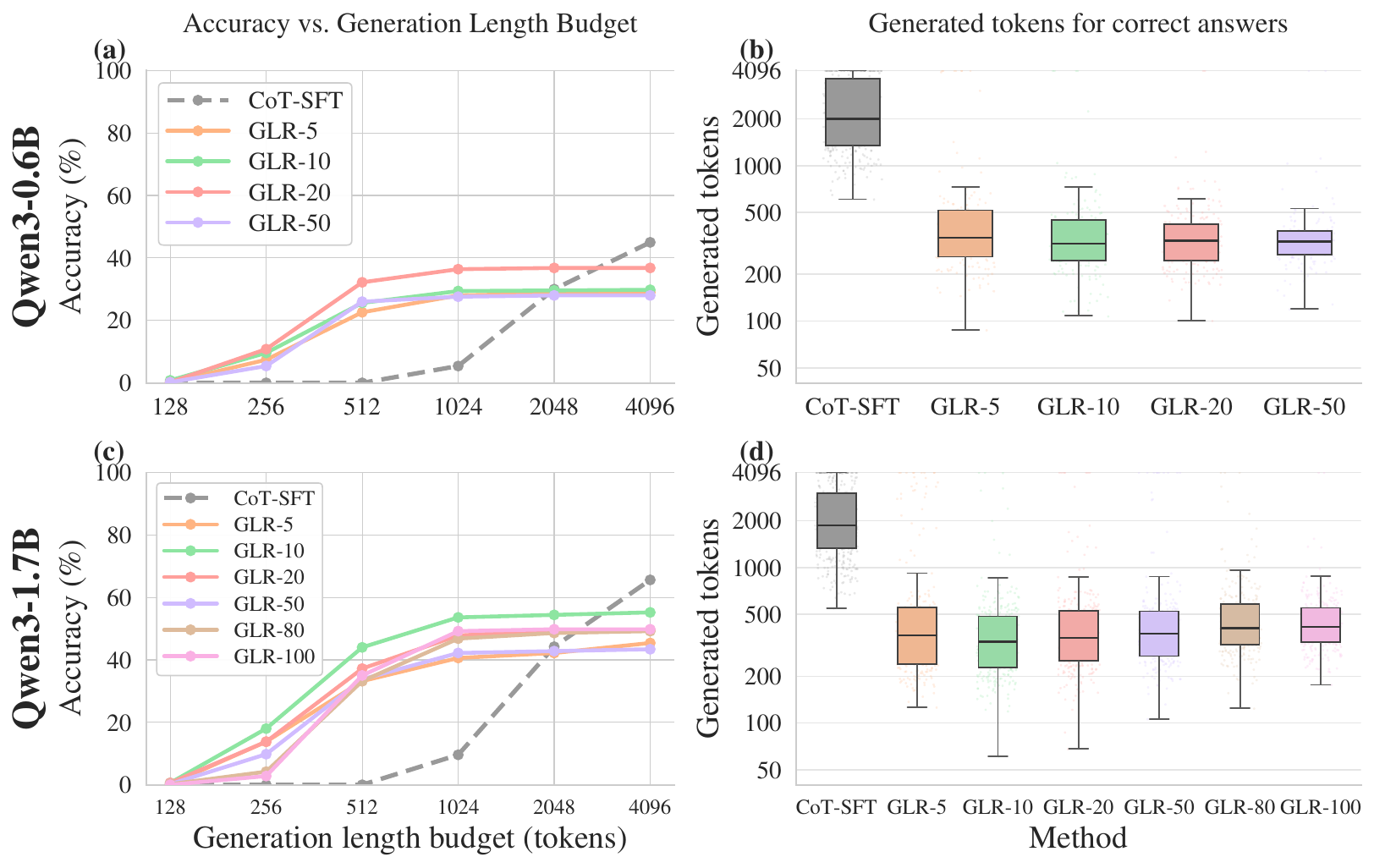}
    \caption{
    \textbf{Emergent reduction in generation length on complex mathematical reasoning (MATH500).} 
    \textbf{Left:} Pass@1 accuracy vs generation budget. \GLR{} shifts the accuracy--length frontier leftward on problems requiring long derivations. 
    \textbf{Right:} Generation length distributions for correct solutions. Moderate latent steps (e.g., \GLRH{10} or \GLRH{20}) greatly reduce the median number of generated steps compared to the \CoTSFT{} baseline. Decoding cap is 4096 tokens.
    }
    \label{fig:math500-qwen3-combined}
    \vspace{-0.4cm}
\end{figure}

\paragraph{Latent steps shift the accuracy--length frontier.}
Figures~\ref{fig:gsm8k-qwen3-combined} and~\ref{fig:math500-qwen3-combined}
(left columns) show pass@1 accuracy under increasing generation budgets.
At small budgets ($\le 256$ steps on GSM8K and $\le 512$ steps on MATH500),
\CoTSFT{} is near zero accuracy, since its explicit CoT trace is usually
truncated before the model reaches an answer. \GLR{} is substantially more
accurate in this regime.
For example, on MATH500 with Qwen3-1.7B at a 512-step
budget, \CoTSFT{} solves nearly $0\%$ of problems, while \GLRH{10} solves over
$40\%$. Since the budget counts both latent steps and subsequent text tokens,
this gain is not due to a larger generation budget.
In fact, this equal-step budget is conservative: during the first $K$ \GLR{}
steps, the model uses the lightweight \thead{} rather than the full vocabulary
projection.
The constrained-budget gain therefore suggests that
the initial latent transitions replace part of the explicit reasoning prefix,
allowing token decoding to resume from a more advanced reasoning state.
This interpretation is supported by the \(K=0\) ablation in
Section~\ref{sec:analysis}. Disabling latent rollout at inference causes the
same \GLR{} model to return to long explicit generations, whereas using even a small number of continuous latent steps substantially shortens successful trajectories.
Thus, the constrained-budget gains arise from the use of continuous transitions at inference, not merely from the \GLR{} training recipe or loss masking.
These findings also hold against an unfrozen-embedding \CoTSFT{} baseline (Appendix~\ref{app:embedding-ablation}). Comparisons with Soft Thinking and a Coconut-style baseline are reported in Appendix~\ref{app:continuous-baselines}.

\paragraph{Successful generations require fewer generated steps.}
The right columns of Figures~\ref{fig:gsm8k-qwen3-combined} and
\ref{fig:math500-qwen3-combined} show total generation length conditioned on
correctness, counting both latent steps and text tokens. Across model sizes and
benchmarks, correct \GLR{} generations require substantially fewer steps than
correct \CoTSFT{} generations. On MATH500 with Qwen3-1.7B, the median correct
\CoTSFT{} generation is approximately 2,000 tokens, whereas moderate latent
budgets such as \GLRH{10} and \GLRH{20} reduce the median to roughly 350 total
steps. This reduction is not explicitly optimized: \GLR{} uses no length penalty
and is trained only to match local embedding-space transitions while preserving
answer generation (Section~\ref{sec:method}). The shorter successful
trajectories therefore suggest that the latent prefix carries part of the
reasoning state that \CoTSFT{} must otherwise externalize through many
autoregressive tokens. Appendix~\ref{app:tok-dist} shows the corresponding
length distributions over all evaluated examples, including incorrect and
truncated generations.
Appendix~\ref{app:greedy-robustness} reports uncertainty estimates for accuracy and length, including comparisons on problems solved by both methods. Appendix~\ref{app:sampling-robustness} further shows that the constrained-budget accuracy advantage persists under temperature sampling.

\paragraph{The latent-step budget controls the accuracy--length tradeoff.}
The latent-step budget $K$ determines how much of the reasoning prefix is
performed in continuous space before token decoding resumes. Its effect is
non-monotonic (Figures~\ref{fig:gsm8k-qwen3-combined} and
\ref{fig:math500-qwen3-combined}): moderate values of $K$ yield the strongest
accuracy--length tradeoff, whereas large values such as \GLRH{80} or
\GLRH{100} on Qwen3-1.7B reduce accuracy. This suggests a stability limit for
uninterrupted latent reasoning. Since $g_\phi$ is trained as a local transition
model, repeatedly applying it without discrete token grounding can accumulate
errors and move the latent state away from the token-induced reasoning
trajectory. We analyze this geometric drift directly in Section~\ref{sec:analysis}.

\paragraph{Latent and explicit reasoning are complementary at large budgets.}
At the largest generation budgets we evaluate (2048 steps for GSM8K and 4096
steps for MATH500), \CoTSFT{} recovers and often outperforms \GLR{} in final
accuracy. Thus, \GLR{} improves the accuracy--length frontier mainly in the
constrained-budget regime. One likely factor is accumulated geometric drift:
\GLR{} replaces an early discrete prefix with autoregressive latent updates, so
local transition errors can move the state away from the token-induced
reasoning trajectory before text decoding resumes. This effect may be amplified
by training scale: due to compute constraints, our \GLR{} models are trained on
only 10K CoT examples, which may not be enough to align the continuous
\thead{} across the full range of token-induced trajectories. These results
suggest a complementary role for the two modes: latent transitions can compress
early reasoning, while explicit tokens provide a more stable scratchpad when
large decoding budgets are available.

Appendix~\ref{app:model-scaling} extends the comparison to Qwen3-4B and Qwen3-8B, where the constrained-budget accuracy advantage and shorter correct-answer generations persist.

\paragraph{Wall-clock inference efficiency.}
To assess whether shorter generations translate into inference savings, we evaluate Qwen3-1.7B \CoTSFT{} and \GLRH{10} using the same Hugging Face decoder, with BF16, FlashAttention-2, KV caching, greedy decoding, and batch size one on NVIDIA H100 MIG 3g.40gb partitions. Across the complete MATH500 and SVAMP test sets, median generation latency decreases from 45.72 to 8.64 seconds and from 10.28 to 2.24 seconds, respectively, corresponding to $5.29\times$ and $4.60\times$ speedups. Timing includes prefill and all latent and textual decoding steps. These measurements show that shorter \GLR{} generations yield wall-clock savings under a matched implementation. Appendix~\ref{app:compute-accounting} provides the full results, measurement protocol, and runtimes of our released vLLM-based~\citep{kwon2023efficient} decoder.

\subsection{Generalization to other benchmarks}
\paragraph{Shorter generations emerge across diverse math benchmarks.}
We next evaluate \GLR{} on four additional benchmarks: SVAMP, MultiArith,
AMC23, and OlympiadBench. These datasets span simple arithmetic, multi-step
word problems, and competition-style mathematics. The same qualitative pattern
appears across these settings: \GLR{} improves accuracy under constrained
generation budgets and reduces the number of generated steps among correct
solutions. Full accuracy--length curves and length distributions for
MultiArith, AMC23, and OlympiadBench are reported in
Appendix~\ref{app:res-other-benchmarks} and \ref{app:tok-dist}.

\begin{figure}[t]
    \centering
    \includegraphics[width=\linewidth]{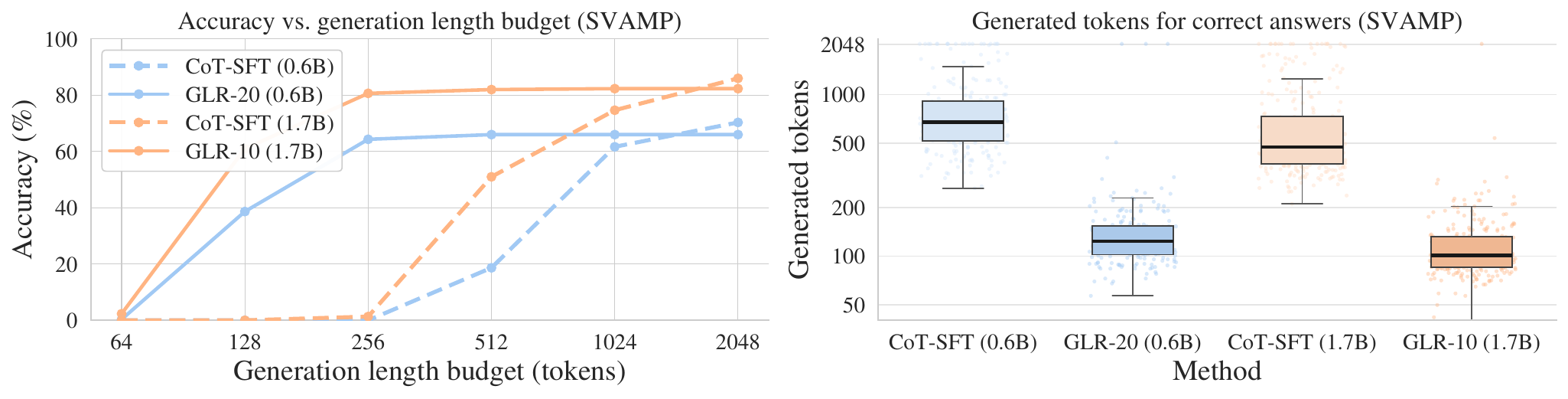} 
    \caption{
    \textbf{\GLR{} reduces redundant reasoning traces on SVAMP.} 
    \textbf{Left:} Accuracy vs. generation budget. On these simpler arithmetic problems where \CoTSFT{} generates long traces, \GLR{} maintains high accuracy under strict budgets ($\le 128$ or $256$ tokens). 
    \textbf{Right:} Generation length distributions for correct answers. While \CoTSFT{} expends hundreds of generated tokens to solve simple problems, \GLR{} reduces the median generation length to approximately 100 steps.
    }
    \label{fig:svamp-qwen3}
    \vspace{-0.4cm}
\end{figure}

\paragraph{Latent transitions bypass redundant steps on simple arithmetic.}
SVAMP provides a direct test of whether shorter generations reflect useful
latent computation rather than only benchmark difficulty
(Figure~\ref{fig:svamp-qwen3}). Most problems require simple arithmetic
operations, such as addition and subtraction. Nevertheless, \CoTSFT{} produces
long explicit reasoning traces: correct solutions have median lengths of
roughly 500--700 tokens depending on model size. This shows that explicit CoT
can incur large serialization overhead even when the required computation is
short. \GLR{} reduces this overhead sharply, solving the same problems in
roughly 100 total generated steps. This gap supports our hypothesis:
the latent prefix carries part of the early reasoning state, allowing the model
to skip redundant explicit steps and resume decoding closer to the
answer-producing part of the trajectory.

\subsection{Understanding Latent Dynamics}
\label{sec:analysis}
\paragraph{Continuous displacements drive generation efficiency.}
By construction, the \thead{} $g_\phi$ predicts continuous embedding-space displacement vectors that need not coincide with exact transitions between vocabulary embeddings. To isolate the effect of this continuous deviation on the reasoning process, we evaluate the Qwen3-1.7B \GLR{} model at $K=0$. In this regime, the model cannot use $g_\phi$ at inference and instead follows exact discrete token updates from the first reasoning step.
\begin{wrapfigure}{r}{0.5\textwidth}
    \vspace{-10pt}
    \centering
    \includegraphics[width=\linewidth]{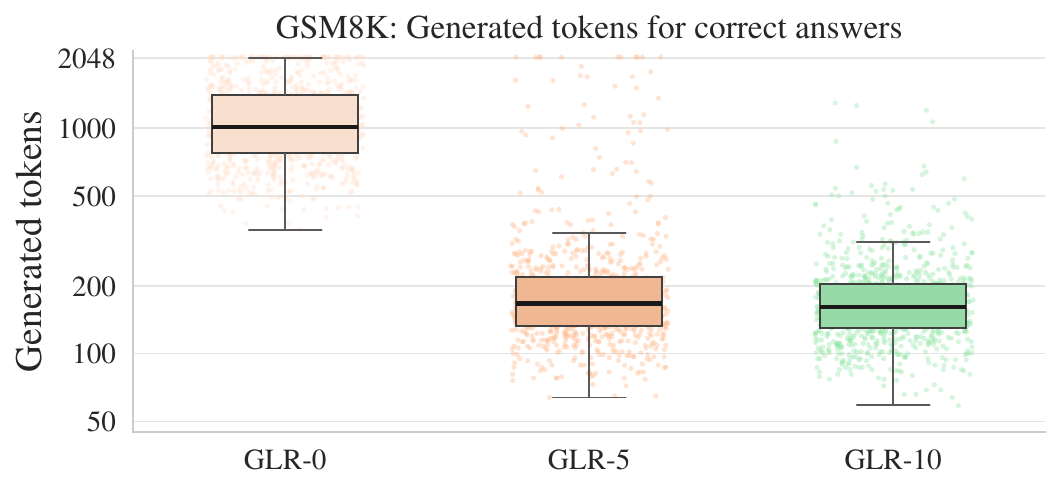} 
    \caption{Generation length for Qwen3-1.7B \GLR{} model at $K=0$ vs. $K>0$ on GSM8K.}
    \label{fig:glr-zero-ablation}
    \vspace{-10pt}
\end{wrapfigure}
Thus, \(K=0\) controls for the training recipe itself: it uses the same \GLR{}-trained backbone and loss masking, but disables latent rollout at inference.
As shown in Figure~\ref{fig:glr-zero-ablation}, this exact discrete update
regime produces long successful trajectories on GSM8K, with a median length of
approximately 1,000 generated tokens. In contrast, using the learned continuous displacements predicted by $g_\phi$ for only a few steps ($K \in \{5, 10\}$) reduces the median length to under 200 tokens. This gap shows that the learned deviations from the
token-embedding path are not arbitrary perturbations. They carry useful
reasoning state, allowing the model to move away from exact token-by-token
transitions and resume decoding closer to an answer-producing region of the
reasoning trajectory.
The large gap between $K=0$ and $K>0$ also indicates that the length reduction is not explained solely by the absence of CE on reasoning tokens; it appears only when the learned continuous transitions are actually used at inference.
Appendix~\ref{app:transition-controls} further examines zero, repeated-marker, random-head, and noisy transitions to assess the role of learned directions.

\paragraph{Continuous representations transition into explicit text.}
We also inspect the text generated immediately after the $K$ latent updates.
Appendix~\ref{app:examples} shows examples where decoding resumes mid-reasoning,
using problem-specific quantities that would normally appear earlier in an
explicit CoT trace. This provides direct qualitative support for our main
interpretation: the latent prefix does more than shorten the visible text; it
moves the model into a partially advanced reasoning state before standard
decoding resumes.

\paragraph{Reasoning progress and embedding-space drift.}
\newsavebox{\latentmetricstablebox}
\sbox{\latentmetricstablebox}{%
    \small
    \setlength{\tabcolsep}{3pt}%
    \begin{tabular}{@{}lccc@{}}
        \toprule
        Benchmark & Progress (\%) & Drift & FH@10 \\
                  & C / I & C / I & (\%) \\
        \midrule
        GSM8K   & 90.5 / 85.7 & .985 / .984 & 0.20 \\
        MATH500 & 81.0 / 70.9 & .989 / .987 & 0.00 \\
        SVAMP   & 89.2 / 85.2 & .987 / .982 & 0.13 \\
        \bottomrule
    \end{tabular}%
}
\begin{wraptable}{R}{\wd\latentmetricstablebox}
    \vspace{-\intextsep}
    \centering
    \captionsetup{width=\linewidth}
    \caption{\textbf{Contextual progress and embedding-space drift for Qwen3-1.7B \GLRH{10}.} C / I: correct / incorrect. Progress and drift are final-step means, reported as a percentage of the reference trace and a cosine distance, respectively; FH@10 is the mean hit rate over the rollout.}
    \label{tab:latent-progress-drift}
    \usebox{\latentmetricstablebox}
\end{wraptable}
To examine how latent states relate to explicit reasoning, we analyze Qwen3-1.7B \GLRH{10} on GSM8K, MATH500, and SVAMP. We align each latent state with a position in the corresponding \CoTSFT{} reasoning trace using contextual hidden-state similarity. Progress is the aligned position divided by the trace length; higher values indicate alignment with later parts of the explicit reasoning. Drift is the cosine distance between the latent input embedding and the aligned CoT token embedding; lower values indicate closer embedding-space correspondence. FutureHit@10 measures the frequency with which a latent state's ten nearest vocabulary tokens match tokens in the reference CoT. We check a 32-token window starting at the aligned position. Table~\ref{tab:latent-progress-drift} shows that correct generations reach higher final progress on all three benchmarks, while drift is similar between correct and incorrect generations. FutureHit@10 is below 0.3\%, indicating that nearest-token projections rarely recover tokens in the aligned window. These findings suggest that correct generations align with later parts of the CoT trace, even though their latent embeddings remain far from the corresponding token embeddings. Appendix~\ref{app:latent-dynamics-protocol} provides the computation details, trajectory curves, and statistical analysis.

\section{Conclusions}
We introduced Geometric Latent Reasoning (\GLR{}), formulating LLM reasoning as a continuous path-approximation problem within the pretrained token-embedding space. By training a lightweight transition head to predict local, CoT-anchored directional updates, \GLR{} shifts computation from discrete text to continuous representations. Our evaluations on mathematical benchmarks demonstrate that replacing early explicit reasoning with these latent steps implicitly induces shorter generations, reaching correct answers using substantially fewer generated tokens. By exposing a controllable inference-time tradeoff between latent transitions, output length, and accuracy, \GLR{} provides a principled geometric foundation for token-efficient reasoning.

\section{Acknowledgements}
Shashi Kumar was partially supported by the EU Horizon 2020 project ELOQUENCE (grant number 101070558).
Yacouba Kaloga was partially supported by Swiss National Science
Foundation project no CRSII5\_202228 on Characterisation of motor speech disorders and processes.

\bibliographystyle{unsrt}
\bibliography{neurips2026/latent_reason}


\appendix
\newpage
\section{Hyperparameters}
\label{app:training-details}
Both the \CoTSFT{} baseline and our proposed \GLR{} models were fine-tuned using the identical underlying language modeling hyperparameters to ensure a strictly controlled comparison. Training was conducted using bfloat16 mixed precision on one Nvidia H100 (80GB) GPU. 
The models were trained for 5 epochs with a cosine learning rate scheduler and a 5\% warmup ratio. To manage memory with a large maximum sequence length of 8,192 tokens, we employed gradient checkpointing and a micro-batch size of 1, utilizing 16 gradient accumulation steps to achieve an effective global batch size of 16. Qwen3-8B uses a 4,096-token training limit and the resulting 5.4K-example subset, as described in Appendix~\ref{app:model-scaling}. Table~\ref{tab:hyperparameters} details the standard optimization hyperparameters.

\begin{table}[h]
    \centering
    \caption{Standard fine-tuning hyperparameters shared across all models.}
    \begin{tabular}{lc}
        \toprule
        \textbf{Hyperparameter} & \textbf{Value} \\
        \midrule
        Training Epochs & 5 \\
        Max Sequence Length & 8,192 \\
        Effective Global Batch Size & 16 \\
        Micro-batch Size & 1 \\
        Gradient Accumulation Steps & 16 \\
        Optimizer & AdamW \\
        Adam $\beta_1, \beta_2$ & 0.9, 0.95 \\
        Peak Learning Rate & $4 \times 10^{-5}$ \\
        Learning Rate Schedule & Cosine \\
        Warmup Ratio & 0.05 \\
        Weight Decay & $1 \times 10^{-4}$ \\
        Precision & \texttt{bfloat16} \\
        Gradient Checkpointing & True \\
        \bottomrule
    \end{tabular}
    \label{tab:hyperparameters}
\end{table}

\paragraph{\GLR{}-Specific Configurations.}
For models equipped with Geometric Latent Reasoning, the token embedding layer was frozen (\texttt{freeze\_input\_embeddings=True}) as discussed in Section~\ref{sec:exp-setup}. The \thead{} was trained using Mean Squared Error (MSE) to predict the directional displacement vectors ($\Delta \ve_i$). The transition objective discount factor ($\gamma$, representing the decay over the reasoning sequence) was set to $0.999$. Finally, aligned with our formulation in Section~\ref{sec:method}, we explicitly disabled cross-entropy supervision on the latent replacement tokens (\texttt{ce\_latent\_tokens=False}) during the second forward pass, ensuring the \thead{} was optimized strictly via the geometric transition objective.

\section{Dataset and Benchmark Details}
\label{app:benchmarks}
This section provides additional details regarding the datasets used for fine-tuning our models and the benchmarks used for evaluation. 

\subsection{Training Data}
\label{app:datasets-train}
All models (\CoTSFT{} and \GLR{}) were trained on a controlled subset of the \textbf{Mixture-of-Thoughts} dataset, provided as part of the Open-R1 initiative \citep{openr1,lozhkov2025openr1math220k}. 
\begin{itemize}
    \item \textbf{Dataset Source:} \url{https://huggingface.co/datasets/open-r1/Mixture-of-Thoughts}
    \item \textbf{Filtering and Processing:} We randomly sampled exactly 10,000 examples from the \texttt{math} split. This dataset provides high-quality, supervised chain-of-thought traces ideal for anchoring our geometric transition objective. To fit the computational limits of our training setup, we filtered out any examples where the total tokenized length (prompt + reasoning trace + answer) exceeded 8,192 tokens. Qwen3-8B uses a 4,096-token training limit and the resulting 5.4K-example subset, as described in Appendix~\ref{app:model-scaling}.
\end{itemize}

\subsection{Evaluation Benchmarks}
\label{app:datasets-eval}
To assess both foundational arithmetic and highly complex mathematical reasoning, we evaluated our models across 6 distinct benchmarks under greedy decoding. The benchmarks are summarized in Table~\ref{tab:benchmarks-summary}.

\begin{table}[h]
    \centering
    \caption{Summary of mathematical reasoning benchmarks used for evaluation.}
    \begin{tabular}{lccc}
        \toprule
        \textbf{Benchmark} & \textbf{Num. Samples} & \textbf{Problem Type} & \textbf{Difficulty Level} \\
        \midrule
        GSM8K \citep{cobbe2021training} & 1,319 & Arithmetic Word Problems & Foundational \\
        SVAMP \citep{patel2021nlp} & 300 & Arithmetic Word Problems & Foundational \\
        MultiArith \citep{roy2015solving} & 180 & Multi-step Arithmetic & Foundational \\
        MATH500 \citep{hendrycks2021measuring} & 500 & Advanced Mathematics & Hard \\
        AMC23 \citep{yang2024qwen2} & 40 & Competition Mathematics & Very Hard \\
        OlympiadBench \citep{he2024olympiadbench} & 674 & Olympiad-level Mathematics & Extremely Hard \\
        \bottomrule
    \end{tabular}
    \label{tab:benchmarks-summary}
\end{table}

\paragraph{Foundational Arithmetic.}
\begin{itemize}
    \item \textbf{GSM8K} \url{https://huggingface.co/datasets/openai/gsm8k}: A standard dataset of high-quality grade-school math word problems requiring 2 to 8 steps of basic arithmetic.
    \item \textbf{SVAMP} \url{https://huggingface.co/datasets/ChilleD/SVAMP}: A challenge set created by applying varying structures to simple word problems. As discussed in Section~\ref{sec:analysis}, SVAMP is particularly useful for observing how models over-generate on simple logic.
    \item \textbf{MultiArith} \url{https://huggingface.co/datasets/ChilleD/MultiArith}: A dataset specifically focused on arithmetic word problems that require multiple reasoning steps to resolve.
\end{itemize}

\paragraph{Advanced and Competition Mathematics.}
\begin{itemize}
    \item \textbf{MATH500} \url{https://huggingface.co/datasets/HuggingFaceH4/MATH-500}: A 500-problem subset of the MATH dataset, encompassing diverse, structurally complex domains including algebra, geometry, and calculus.
    \item \textbf{AMC23} \url{https://huggingface.co/datasets/math-ai/amc23}: A collection of recent problems from the American Mathematics Competitions (AMC), testing advanced logical deduction and theorem application.
    \item \textbf{OlympiadBench} \url{https://huggingface.co/datasets/Hothan/OlympiadBench/viewer/OE_TO_maths_en_COMP}: The math-specific subset of OlympiadBench, representing the highest tier of problem difficulty (e.g., IMO-level problems) requiring rigorous, long-form derivations. 
\end{itemize}

\section{Results on other benchmarks}
\label{app:res-other-benchmarks}
\begin{figure}[H]
    \centering
    \includegraphics[width=\linewidth]{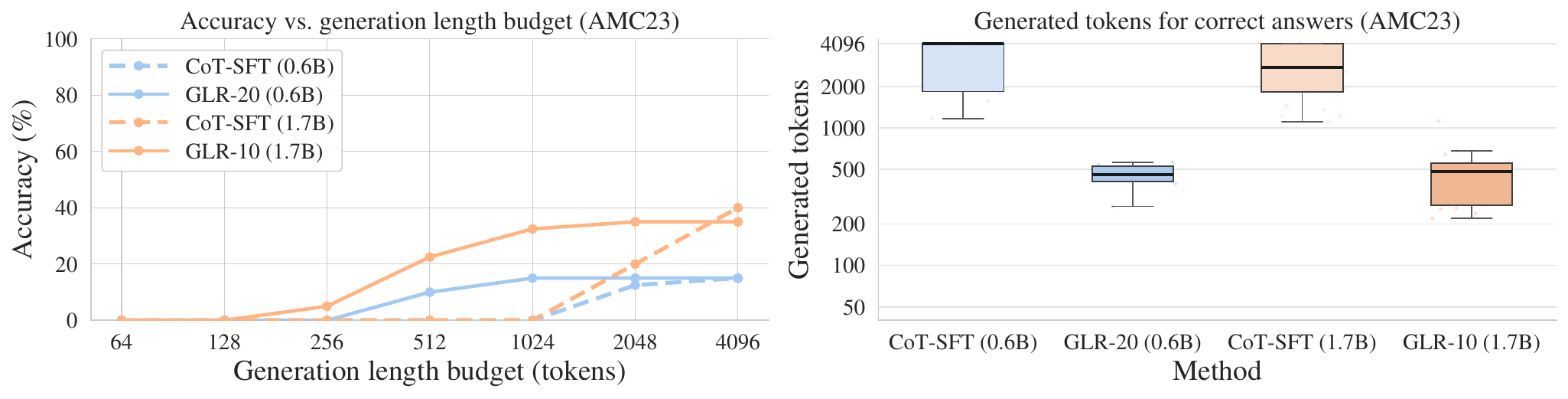}
    \caption{\textbf{Accuracy and generation length on AMC23.} \GLR{} reduces the median generation length on competition mathematics).
    }
    \label{fig:amc23-qwen3-combined}
\end{figure}

\begin{figure}[H]
    \centering
    \includegraphics[width=\linewidth]{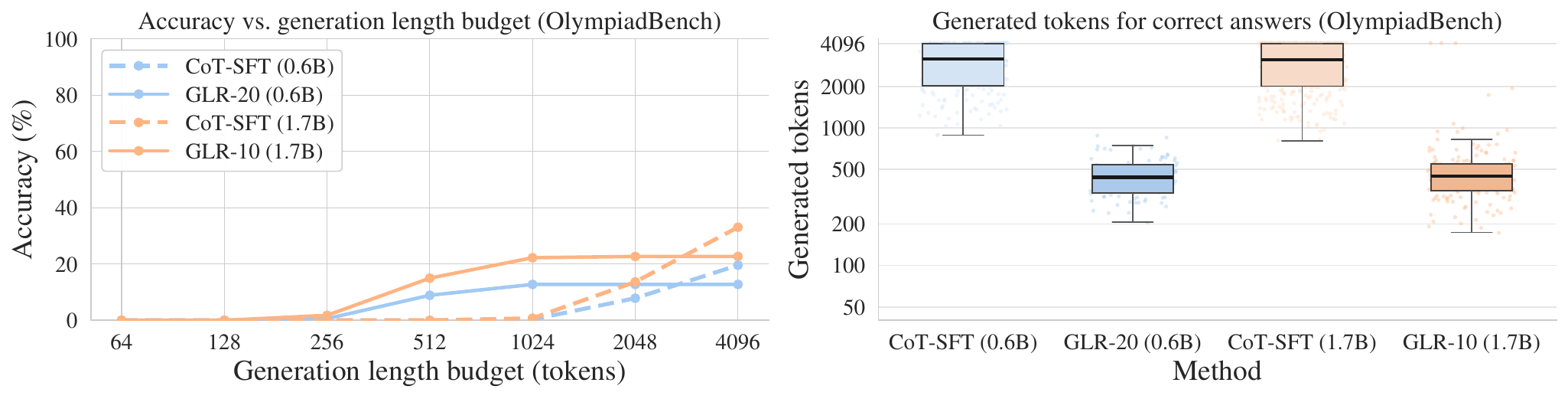}
    \caption{\textbf{Accuracy and generation length on OlympiadBench.} \GLR{} shifts the accuracy--length frontier on Olympiad-level mathematical reasoning.
    }
    \label{fig:ob-qwen3-combined}
\end{figure}

\begin{figure}[H]
    \centering
    \includegraphics[width=\linewidth]{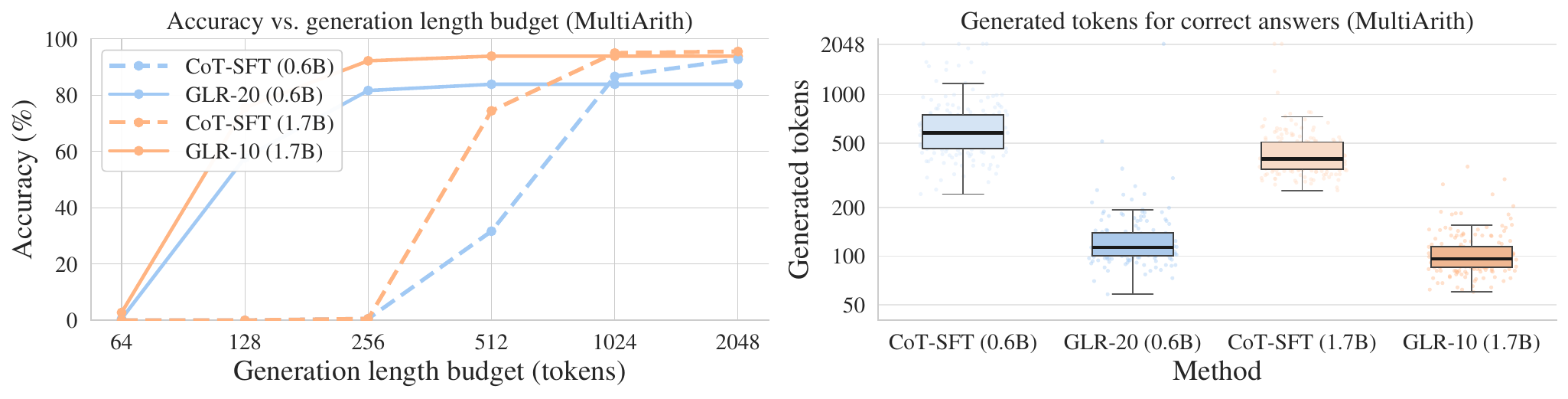}
    \caption{\textbf{Accuracy and generation length on MultiArith.} \GLR{} reduces the generated steps required for foundational multi-step arithmetic problems.
    }
    \label{fig:multiarith-qwen3-combined}
\end{figure}

This section provides the full accuracy--length tradeoff curves and generation length distributions (for correct answers) on the remaining three evaluation benchmarks. Figures~\ref{fig:amc23-qwen3-combined}, \ref{fig:ob-qwen3-combined}, and \ref{fig:multiarith-qwen3-combined} display the results for AMC23, OlympiadBench, and MultiArith, respectively. 
Consistent with the main findings on GSM8K, MATH500, and SVAMP, \GLR{} shifts the accuracy--length frontier and reduces the median number of generated steps required to solve the problems. This confirms that the emergent reduction in generation length generalizes across both foundational multi-step arithmetic and advanced competition mathematics.

\section{Effect of Unfreezing the Input Embeddings}
\label{app:embedding-ablation}

\paragraph{A matched embedding ablation.}
The primary comparison freezes the input token embeddings for both \GLR{} and \CoTSFT{}, keeping the pretrained embedding geometry fixed. Standard \CoTSFT{} does not require this constraint. We therefore train an additional Qwen3-1.7B \CoTSFT{} baseline with unfrozen input embeddings, using the same training data, and compare it with the frozen baseline and \GLR{}-10. Table~\ref{tab:unfrozen-embedding} separates accuracy at constrained generation budgets from median generation length over correct answers at the full evaluation caps. The latent-step budget is $K=10$ on all five benchmarks.

\begin{table}[htbp]
\centering
\small
\caption{\textbf{Effect of unfreezing the input embeddings for Qwen3-1.7B.} Left: accuracy at the indicated total-step budget. Right: median total steps over correct answers at the benchmark's full evaluation cap. Frozen and unfrozen refer to the input embedding layer; GLR uses frozen input embeddings and $K=10$ latent steps. OB denotes OlympiadBench. Best results are bold.}
\vspace{1mm}
\label{tab:unfrozen-embedding}
\begin{minipage}[t]{0.49\linewidth}
\centering
\small
\setlength{\tabcolsep}{2pt}
\renewcommand{\arraystretch}{1.2}
\textbf{(a) Accuracy (\%)}\par\smallskip
\begin{tabular*}{\linewidth}{@{\extracolsep{\fill}}lrrrr@{}}
\toprule
Benchmark & Budget & \shortstack{CoT-SFT\\(frozen)} & \shortstack{CoT-SFT\\(unfrozen)} & GLR-10 \\
\midrule
\multirow{2}{*}{GSM8K} & 512 & 17.2 & 17.4 & \textbf{70.4} \\
& 1024 & 65.7 & 64.8 & \textbf{71.4} \\
\midrule
\multirow{2}{*}{MATH500} & 1024 & 9.6 & 6.6 & \textbf{53.6} \\
& 2048 & 44.2 & 41.8 & \textbf{54.4} \\
\midrule
SVAMP & 1024 & 74.7 & 74.7 & \textbf{82.3} \\
\midrule
AMC23 & 2048 & 20.0 & 17.5 & \textbf{35.0} \\
\midrule
OB & 2048 & 13.6 & 13.6 & \textbf{22.7} \\
\bottomrule
\end{tabular*}
\end{minipage}\hfill
\begin{minipage}[t]{0.49\linewidth}
\centering
\small
\setlength{\tabcolsep}{2pt}
\renewcommand{\arraystretch}{1.2}
\textbf{(b) Median generation steps}\par\smallskip
\begin{tabular*}{\linewidth}{@{\extracolsep{\fill}}lrrrr@{}}
\toprule
Benchmark & \shortstack{Max\\Budget} & \shortstack{CoT-SFT\\(frozen)} & \shortstack{CoT-SFT\\(unfrozen)} & GLR-10 \\
\midrule
GSM8K & 2048 & 699 & 686 & \textbf{161} \\
\midrule
MATH500 & 4096 & 1876 & 1815 & \textbf{334} \\
\midrule
SVAMP & 2048 & 471 & 462 & \textbf{101} \\
\midrule
AMC23 & 4096 & 2751 & 2895 & \textbf{483} \\
\midrule
OB & 4096 & 3129 & 2599 & \textbf{447} \\
\bottomrule
\end{tabular*}
\end{minipage}
\end{table}

\paragraph{The accuracy and length trends persist with trainable embeddings.}
Unfreezing the input embeddings produces small changes in constrained-budget accuracy and does not close the gap to \GLR{}. For example, GSM8K accuracy at 1,024 steps changes from 65.7\% to 64.8\%, compared with 71.4\% for \GLR{}. Median generation length remains substantially higher for both \CoTSFT{} variants across all five benchmarks. Even on OlympiadBench, where unfreezing reduces the median from 3,129 to 2,599 steps, \GLR{} requires 447 steps. These results indicate that the observed advantage of \GLR{} does not depend on freezing the baseline's input embeddings.

\section{Continuous-Reasoning Baselines}
\label{app:continuous-baselines}

\paragraph{Soft-token interpolation and hidden-state feedback.}
We compare \GLR{} with Soft Thinking~\citep{zhang2025soft} and a Coconut-style hidden-state-feedback baseline~\citep{hao2024training}, using the same model family, benchmark prompts, total-step accounting, and training data where training is required. Soft Thinking performs continuous soft-token reasoning using the \CoTSFT{} checkpoint. We evaluate both fixed soft-token budgets ($K=20$ for 0.6B and $K=10$ for 1.7B) and continuation until the reasoning span closes, with a safety cap of 256 soft steps. We report the variant with stronger budgeted accuracy for each model and benchmark. The Coconut-style baseline uses a two-pass training procedure: a teacher-forced pass constructs hidden-state replacements, and a second pass applies these replacements with answer-token supervision. Inference feeds hidden states back recursively.

\begin{table}[htbp]
\centering
\small
\setlength{\tabcolsep}{4pt}
\renewcommand{\arraystretch}{1.18}
\caption{\textbf{Accuracy of continuous-reasoning baselines.} Accuracy (\%) is measured at the indicated total-step budget. GLR uses $K=20$ at 0.6B and $K=10$ at 1.7B. Soft Thinking uses the stronger evaluated variant for each model and benchmark. Coconut-style denotes the adapted two-pass baseline. Best results within each model size and benchmark are bold.}
\vspace{1mm}
\label{tab:continuous-baselines}
\begin{tabular*}{\linewidth}{@{\extracolsep{\fill}}lccccc@{}}
\toprule
Benchmark & Budget & CoT-SFT & Soft Thinking & Coconut-style & GLR \\
\midrule
\multicolumn{6}{@{}l}{\textbf{Qwen3-0.6B}} \\
\midrule
\multirow{2}{*}{GSM8K} & 512 & 5.2 & 5.1 & 5.5 & \textbf{47.5} \\
 & 1024 & 38.9 & 38.8 & 5.7 & \textbf{47.8} \\
\midrule
\multirow{2}{*}{MATH500} & 1024 & 5.4 & 7.8 & 8.3 & \textbf{36.4} \\
 & 2048 & 30.0 & 33.3 & 8.3 & \textbf{36.8} \\
\midrule
SVAMP & 1024 & 61.7 & 60.0 & 14.3 & \textbf{66.0} \\
\midrule
\multicolumn{6}{@{}l}{\textbf{Qwen3-1.7B}} \\
\midrule
\multirow{2}{*}{GSM8K} & 512 & 17.2 & 17.9 & \textemdash & \textbf{70.4} \\
 & 1024 & 65.7 & 66.4 & \textemdash & \textbf{71.4} \\
\midrule
\multirow{2}{*}{MATH500} & 1024 & 9.6 & 14.6 & \textemdash & \textbf{53.6} \\
 & 2048 & 44.2 & 50.0 & \textemdash & \textbf{54.4} \\
\midrule
SVAMP & 1024 & 74.7 & 79.3 & \textemdash & \textbf{82.3} \\
\bottomrule
\end{tabular*}
\end{table}

\begin{table}[htbp]
\centering
\small
\setlength{\tabcolsep}{4pt}
\renewcommand{\arraystretch}{1.18}
\caption{\textbf{Generation length of continuous-reasoning baselines.} Values are median total steps over correct answers at the indicated maximum generation budget. GLR uses $K=20$ at 0.6B and $K=10$ at 1.7B. Soft Thinking uses the stronger evaluated variant for each model and benchmark. Coconut-style denotes the adapted two-pass baseline. Best results within each model size and benchmark are bold.}
\vspace{1mm}
\label{tab:continuous-baselines-length}
\begin{tabular*}{\linewidth}{@{\extracolsep{\fill}}lccccc@{}}
\toprule
Benchmark & Max Budget & CoT-SFT & Soft Thinking & Coconut-style & GLR \\
\midrule
\multicolumn{6}{@{}l}{\textbf{Qwen3-0.6B}} \\
\midrule
GSM8K & 2048 & 934 & 902 & 655 & \textbf{168} \\
\midrule
MATH500 & 4096 & 1996 & 1760 & 1107 & \textbf{328} \\
\midrule
SVAMP & 2048 & 675 & 662 & 370 & \textbf{123} \\
\midrule
\multicolumn{6}{@{}l}{\textbf{Qwen3-1.7B}} \\
\midrule
GSM8K & 2048 & 699 & 697 & \textemdash & \textbf{161} \\
\midrule
MATH500 & 4096 & 1876 & 1442 & \textemdash & \textbf{334} \\
\midrule
SVAMP & 2048 & 471 & 456 & \textemdash & \textbf{101} \\
\bottomrule
\end{tabular*}
\end{table}

\paragraph{Continuous representations alone do not recover the same tradeoff.}
Table~\ref{tab:continuous-baselines} shows that Soft Thinking improves some constrained-budget results, but \GLR{} retains the highest accuracy across the reported settings. On Qwen3-1.7B MATH500, Soft Thinking raises accuracy at 2,048 steps from 44.2\% to 50.0\%, compared with 54.4\% for \GLR{}-10. Table~\ref{tab:continuous-baselines-length} separately compares generation length at the full evaluation caps. On the same benchmark, the median decreases from 1,876 steps for \CoTSFT{} to 1,442 for Soft Thinking and 334 for \GLR{}-10. The Coconut-style baseline also produces shorter generations than \CoTSFT{}, but its accuracy is substantially lower at the larger reported budgets. Together, the two comparisons show that \GLR{} combines higher constrained-budget accuracy with shorter correct-answer generations, supporting the value of learned embedding-space transitions within the evaluated procedures.

\section{Scaling to Larger Models}
\label{app:model-scaling}

\paragraph{Matched comparisons at 4B and 8B.}
We extend the evaluation to Qwen3-4B and Qwen3-8B. Within each model size, \CoTSFT{} and \GLR{} use the same training data and optimization configuration. Qwen3-4B follows the 10K-example training setup used for the smaller models. For Qwen3-8B, the memory requirements of full-parameter training on a single H100 require filtering sequences longer than 4,096 tokens, leaving 5.4K training examples. Both 8B methods use this same subset. Thus, the comparison between methods remains matched within each scale, while comparisons across scales also reflect the different training subset at 8B.

\begin{table}[ht]
\centering
\setlength{\tabcolsep}{4pt}
\renewcommand{\arraystretch}{1.18}
\caption{\textbf{Accuracy across model scales.} Accuracy (\%) is measured at the indicated total-step budget. Size denotes the Qwen3 model size. Bold marks the better result between CoT-SFT and GLR within each model size and benchmark. Both 8B methods use the same 5.4K-example training subset with a 4,096-token training limit.}
\vspace{1mm}
\label{tab:model-scaling}
\begin{tabular*}{\linewidth}{@{\extracolsep{\fill}}lcccc@{}}
\toprule
Benchmark & Size & Budget & CoT-SFT & GLR \\
\midrule
\multirow{8}{*}{GSM8K} & \multirow{2}{*}{0.6B} & 512 & 5.2 & \textbf{47.5} \\
 &  & 1024 & 38.9 & \textbf{47.8} \\
\addlinespace[2pt]
 & \multirow{2}{*}{1.7B} & 512 & 17.2 & \textbf{70.4} \\
 &  & 1024 & 65.7 & \textbf{71.4} \\
\addlinespace[2pt]
 & \multirow{2}{*}{4B} & 512 & 14.9 & \textbf{79.2} \\
 &  & 1024 & 68.7 & \textbf{81.2} \\
\addlinespace[2pt]
 & \multirow{2}{*}{8B} & 512 & 15.5 & \textbf{81.6} \\
 &  & 1024 & 70.1 & \textbf{88.1} \\
\midrule
\multirow{8}{*}{MATH500} & \multirow{2}{*}{0.6B} & 1024 & 5.4 & \textbf{36.4} \\
 &  & 2048 & 30.0 & \textbf{36.8} \\
\addlinespace[2pt]
 & \multirow{2}{*}{1.7B} & 1024 & 9.6 & \textbf{53.6} \\
 &  & 2048 & 44.2 & \textbf{54.4} \\
\addlinespace[2pt]
 & \multirow{2}{*}{4B} & 1024 & 10.0 & \textbf{56.4} \\
 &  & 2048 & 49.0 & \textbf{58.0} \\
\addlinespace[2pt]
 & \multirow{2}{*}{8B} & 1024 & 7.6 & \textbf{66.4} \\
 &  & 2048 & 47.2 & \textbf{67.8} \\
\midrule
\multirow{4}{*}{SVAMP} & 0.6B & 1024 & 61.7 & \textbf{66.0} \\
 & 1.7B & 1024 & 74.7 & \textbf{82.3} \\
 & 4B & 1024 & 75.0 & \textbf{86.0} \\
 & 8B & 1024 & 76.7 & \textbf{89.3} \\
\midrule
\multirow{4}{*}{AMC23} & 0.6B & 2048 & 12.5 & \textbf{15.0} \\
 & 1.7B & 2048 & 20.0 & \textbf{35.0} \\
 & 4B & 2048 & 22.5 & \textbf{40.0} \\
 & 8B & 2048 & 22.5 & \textbf{70.0} \\
\bottomrule
\end{tabular*}
\end{table}

\begin{table}[ht]
\centering
\setlength{\tabcolsep}{4pt}
\renewcommand{\arraystretch}{1.18}
\caption{\textbf{Generation length across model scales.} Values are median total steps over correct answers at the indicated maximum generation budget. Size denotes the Qwen3 model size. Bold marks the better result between CoT-SFT and GLR within each model size and benchmark. Both 8B methods use the same 5.4K-example training subset with a 4,096-token training limit.}
\label{tab:model-scaling-length}
\begin{tabular*}{\linewidth}{@{\extracolsep{\fill}}lcccc@{}}
\toprule
Benchmark & Size & Max Budget & CoT-SFT & GLR \\
\midrule
\multirow{4}{*}{GSM8K} & 0.6B & 2048 & 934 & \textbf{168} \\
 & 1.7B & 2048 & 699 & \textbf{161} \\
 & 4B & 2048 & 728 & \textbf{252} \\
 & 8B & 2048 & 748 & \textbf{228} \\
\midrule
\multirow{4}{*}{MATH500} & 0.6B & 4096 & 1996 & \textbf{328} \\
 & 1.7B & 4096 & 1876 & \textbf{334} \\
 & 4B & 4096 & 1816 & \textbf{374} \\
 & 8B & 4096 & 1865 & \textbf{413} \\
\midrule
\multirow{4}{*}{SVAMP} & 0.6B & 2048 & 675 & \textbf{123} \\
 & 1.7B & 2048 & 471 & \textbf{101} \\
 & 4B & 2048 & 576 & \textbf{124} \\
 & 8B & 2048 & 568 & \textbf{147} \\
\midrule
\multirow{4}{*}{AMC23} & 0.6B & 4096 & 4092 & \textbf{459} \\
 & 1.7B & 4096 & 2751 & \textbf{483} \\
 & 4B & 4096 & 2885 & \textbf{472} \\
 & 8B & 4096 & 3424 & \textbf{486} \\
\bottomrule
\end{tabular*}
\end{table}

\paragraph{The accuracy and length advantage extends to larger models.}
Table~\ref{tab:model-scaling} shows that the constrained-budget accuracy advantage persists at both 4B and 8B. At 8B, \GLR{} improves GSM8K accuracy at 512 steps from 15.5\% to 81.6\%, MATH500 accuracy at 1,024 steps from 7.6\% to 66.4\%, and AMC23 accuracy at 2,048 steps from 22.5\% to 70.0\%. The advantage remains at the larger reported budgets. Table~\ref{tab:model-scaling-length} shows that correct-answer generations also remain shorter when evaluated at the full generation caps. The 8B median decreases from 748 to 228 steps on GSM8K and from 3,424 to 486 on AMC23, with reductions between 69.5\% and 85.8\% across the four benchmarks. These results extend the evidence for token-efficient reasoning beyond the smaller models used in the main analysis, without an explicit length objective.

\section{Generated token distribution over full benchmark}
\label{app:tok-dist}
\begin{figure}[htbp]
    \centering
    \includegraphics[width=\linewidth]{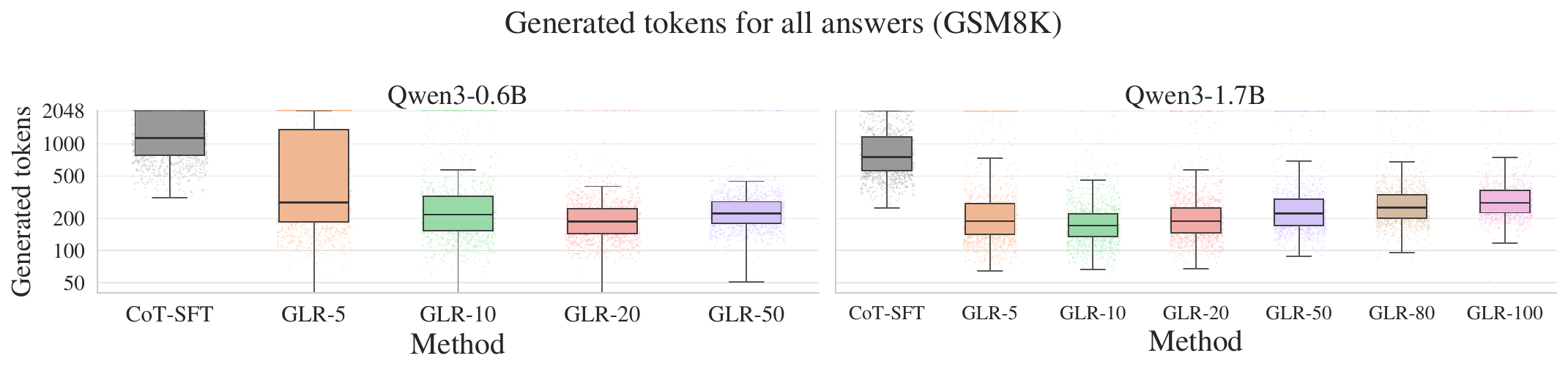}
    \caption{Distribution of total generated tokens across all answers on GSM8K.
    }
    \label{fig:gsm8k-qwen3-combined-full}
\end{figure}

\begin{figure}[htbp]
    \centering
    \includegraphics[width=\linewidth]{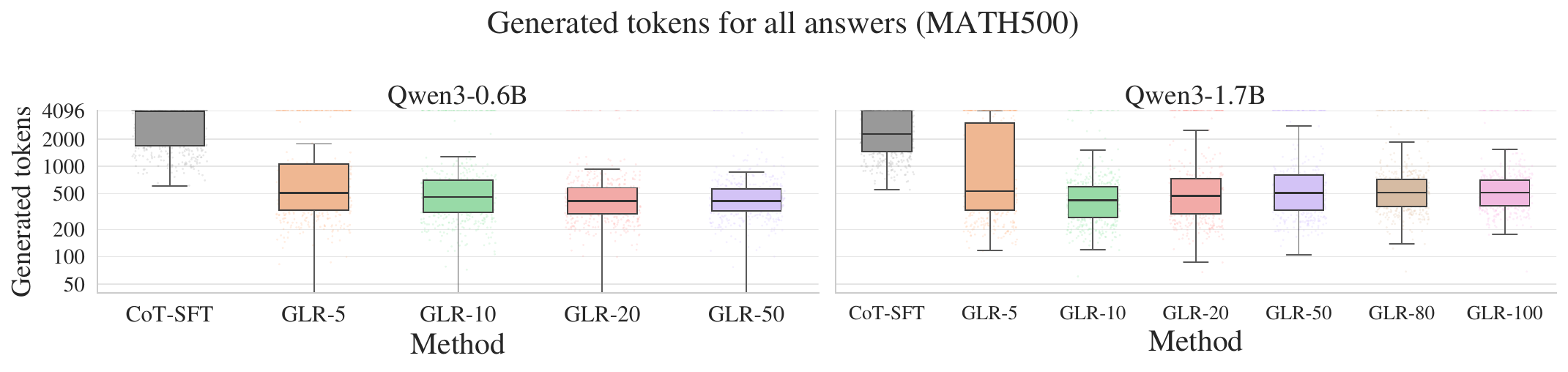}
    \caption{Distribution of total generated tokens across all answers on MATH500.
    }
    \label{fig:math500-qwen3-combined-full}
\end{figure}

\begin{figure}[htbp]
    \centering
    \includegraphics[width=\linewidth]{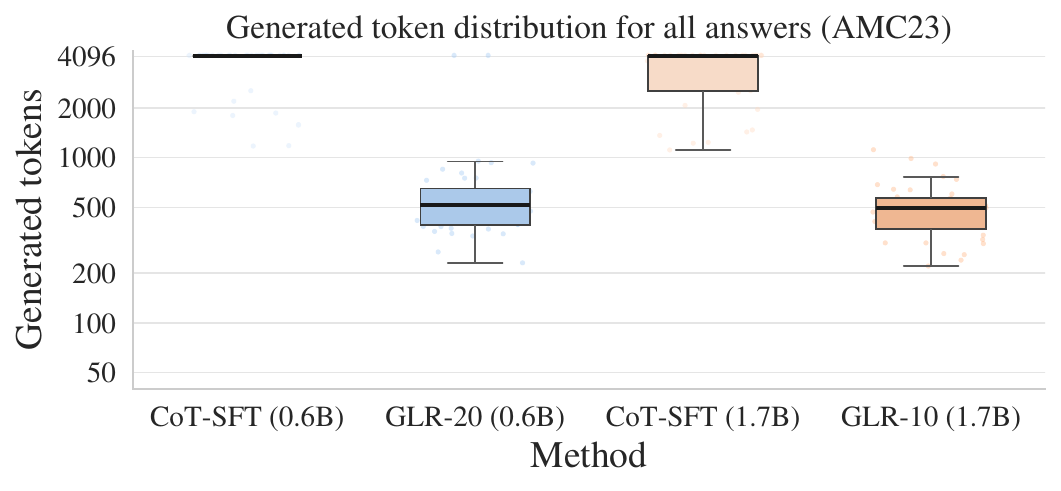}
    \caption{Distribution of total generated tokens across all answers on AMC23.
    }
    \label{fig:amc23-qwen3-combined-full}
\end{figure}

\begin{figure}[htbp]
    \centering
    \includegraphics[width=\linewidth]{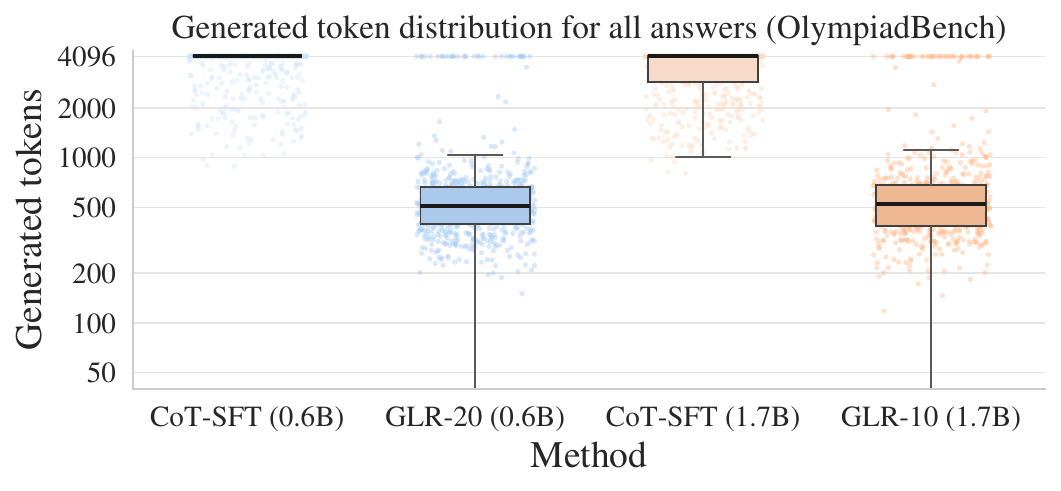}
    \caption{Distribution of total generated tokens across all answers on OlympiadBench.
    }
    \label{fig:ob-qwen3-combined-full}
\end{figure}

\begin{figure}[htbp]
    \centering
    \includegraphics[width=\linewidth]{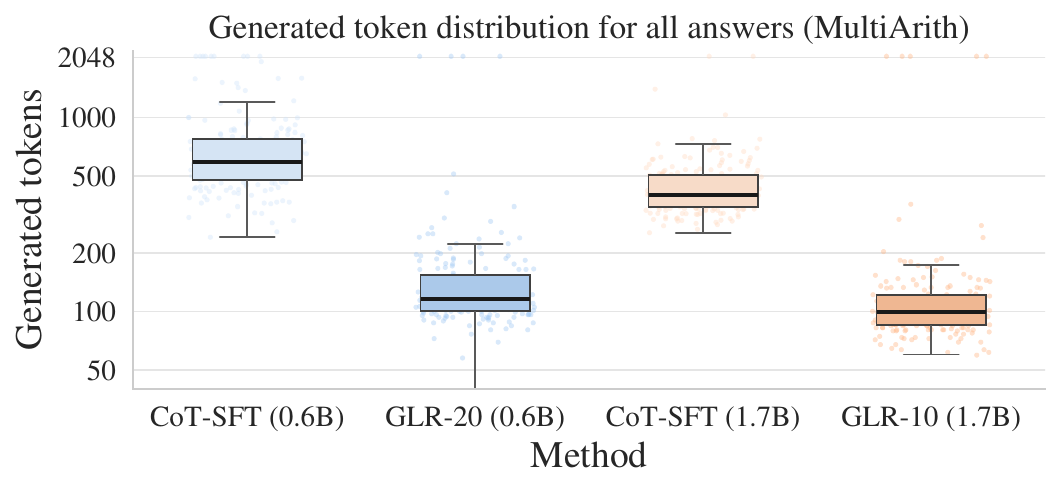}
    \caption{Distribution of total generated tokens across all answers on MultiArith.
    }
    \label{fig:multiarith-qwen3-combined-full}
\end{figure}

In the main text (Section~\ref{sec:res-disc}), we report generation lengths exclusively for correct answers to isolate the number of model steps required for a successful solution. In this section, we present the generation length distributions across \textit{all} evaluated problems, including incorrect answers and generations that reached the decoding cap. 
Figures~\ref{fig:gsm8k-qwen3-combined-full} and \ref{fig:math500-qwen3-combined-full} show these full distributions for the primary benchmarks, while Figures~\ref{fig:amc23-qwen3-combined-full}, \ref{fig:ob-qwen3-combined-full}, and \ref{fig:multiarith-qwen3-combined-full} provide the distributions for the remaining datasets. These figures demonstrate that \GLR{} produces shorter generations across the entire evaluation set. This indicates that the method reduces the overall test-time generation length, rather than only producing shorter outputs when the model successfully reaches the correct answer.


\section{Statistical Robustness of Accuracy and Generation Length}
\label{app:accuracy-length-robustness}

We evaluate the robustness of the accuracy and length comparisons using the fixed Qwen3-1.7B checkpoints, with $K=10$ for \GLR{}. The analysis uses all 1,319 GSM8K, 500 MATH500, and 300 SVAMP test problems. We compute 95\% percentile confidence intervals using 10,000 paired bootstrap resamples of problems within each benchmark, keeping the two methods' outputs for each problem together. These intervals quantify uncertainty over evaluation problems for the trained checkpoints. Accuracy is measured under a fixed generation budget: an output is counted as correct only if it is fully correct and its total generation length does not exceed the budget. Latent positions are counted toward this length.

\subsection{Greedy Accuracy and Generation Length}
\label{app:greedy-robustness}

\begin{table}[htbp]
\centering
\setlength{\tabcolsep}{4pt}
\renewcommand{\arraystretch}{1.25}
\caption{\textbf{Uncertainty in constrained-budget accuracy for Qwen3-1.7B.} Accuracy is reported as a percentage; $\Delta$ is GLR-10 minus CoT-SFT in percentage points. Bracketed values below each estimate give 95\% paired-bootstrap confidence intervals. Bold marks the higher accuracy. Exact two-sided McNemar tests give $p_{\mathrm{Holm}}<0.001$ on each benchmark after correction across the three comparisons.}
\label{tab:greedy-accuracy-robustness}
\begin{tabular*}{\linewidth}{@{\extracolsep{\fill}}lcccc@{}}
\toprule
Benchmark & Budget & \shortstack{CoT-SFT\\Acc. (\%)} & \shortstack{GLR-10\\Acc. (\%)} & $\Delta$ (pp) \\
\midrule
GSM8K & 256 & \shortstack{0.1\\{}[0.0, 0.2]} & \shortstack{\textbf{63.7}\\{}[61.1, 66.3]} & \shortstack{63.6\\{}[61.0, 66.2]} \\
\midrule
MATH500 & 512 & \shortstack{0.0\\{}[0.0, 0.0]} & \shortstack{\textbf{44.0}\\{}[39.6, 48.4]} & \shortstack{44.0\\{}[39.6, 48.4]} \\
\midrule
SVAMP & 256 & \shortstack{1.3\\{}[0.3, 2.7]} & \shortstack{\textbf{80.7}\\{}[76.0, 85.0]} & \shortstack{79.3\\{}[74.7, 84.0]} \\
\bottomrule
\end{tabular*}
\end{table}

\paragraph{The constrained-budget accuracy differences exceed sampling uncertainty.}
At the representative budgets in Table~\ref{tab:greedy-accuracy-robustness}, every paired confidence interval for the accuracy difference lies above zero. The numbers of problems solved only by \CoTSFT{} and only by \GLR{} are respectively $(0,839)$ on GSM8K, $(0,220)$ on MATH500, and $(1,239)$ on SVAMP. Exact two-sided McNemar tests with Holm correction across the three benchmarks give adjusted $p<0.001$ in each case. This supports the constrained-budget accuracy advantage, including on the smaller SVAMP test set.

\begin{table}[htbp]
\centering
\setlength{\tabcolsep}{4pt}
\renewcommand{\arraystretch}{1.25}
\caption{\textbf{Uncertainty in generation-length reductions for Qwen3-1.7B.} Median lengths count all generation steps; $n$ is the number of problems in each subset. The upper panel includes incorrect and capped generations; the lower panel uses problems solved by both methods. Reduction is $100(1-m_{\mathrm{GLR}}/m_{\mathrm{CoT}})$, where $m$ is the subset median. Bracketed values below estimates give 95\% paired-bootstrap confidence intervals. Bold marks the lower median length.}
\label{tab:length-robustness}
\begin{tabular*}{\linewidth}{@{\extracolsep{\fill}}lccccc@{}}
\toprule
Benchmark & \shortstack{Max\\Budget} & $n$ & \shortstack{CoT-SFT\\Median steps} & \shortstack{GLR-10\\Median steps} & \shortstack{Reduction\\(\%)} \\
\midrule
\multicolumn{6}{@{}l}{\textbf{All evaluation problems}} \\
\midrule
GSM8K & 2048 & 1319 & \shortstack{753\\{}[730, 778]} & \shortstack{\textbf{172}\\{}[168, 176]} & \shortstack{77.2\\{}[76.3, 77.9]} \\
\midrule
MATH500 & 4096 & 500 & \shortstack{2264.5\\{}[2018, 2481.5]} & \shortstack{\textbf{420.5}\\{}[386.5, 446.5]} & \shortstack{81.4\\{}[79.6, 83.1]} \\
\midrule
SVAMP & 2048 & 300 & \shortstack{493\\{}[460, 541]} & \shortstack{\textbf{108}\\{}[102, 112.5]} & \shortstack{78.1\\{}[76.6, 80.1]} \\
\midrule
\multicolumn{6}{@{}l}{\textbf{Problems solved by both methods}} \\
\midrule
GSM8K & 2048 & 884 & 661.5 & \textbf{159.0} & \shortstack{76.0\\{}[74.9, 76.8]} \\
\midrule
MATH500 & 4096 & 261 & 1596.0 & \textbf{322.0} & \shortstack{79.8\\{}[77.5, 81.2]} \\
\midrule
SVAMP & 2048 & 236 & 453.5 & \textbf{101.0} & \shortstack{77.7\\{}[76.1, 79.0]} \\
\bottomrule
\end{tabular*}
\end{table}

\paragraph{Length reductions persist on the same solved problems.}
To separate shorter generations from differences in which problems each method solves, we analyze both all evaluation problems and the subset solved by both methods. Lengths are capped at 2,048 steps for GSM8K and SVAMP and 4,096 for MATH500. For each bootstrap resample, we recompute the two medians and their relative reduction. Table~\ref{tab:length-robustness} shows median reductions of 77.2\%, 81.4\%, and 78.1\% over all problems on GSM8K, MATH500, and SVAMP, respectively. Restricting the comparison to the same solved problems gives reductions of 76.0\%, 79.8\%, and 77.7\%, with all confidence intervals above zero. Thus, the length advantage persists when both methods solve the same problems and when incorrect or capped generations are included.

\subsection{Robustness to Temperature Sampling}
\label{app:sampling-robustness}

\paragraph{Sampling protocol and answer aggregation.}
For each problem, we generate three trajectories per method with temperature $T=0.6$, top-$p=0.95$, and random seeds 17, 29, and 43. We retain the Qwen3-1.7B checkpoints, \GLR{} budget $K=10$, and constrained total-step budgets from the greedy comparison: 256 on GSM8K and SVAMP and 512 on MATH500. Sampled pass@1 averages correctness over the three trajectories. Majority@3 requires at least two numerically or symbolically equivalent completed answers and counts the resulting consensus as correct only when it matches the reference answer. Incomplete generations abstain from voting. Fewer than two equivalent completed answers yields no consensus and is counted as incorrect. Pass@3 reports whether any of the three trajectories is correct and is an oracle metric. We compute 95\% confidence intervals from 10,000 paired problem-level cluster-bootstrap resamples, retaining all three trajectories from each sampled problem together for both methods.

\begin{table}[htbp]
\centering
\setlength{\tabcolsep}{4pt}
\renewcommand{\arraystretch}{1.25}
\caption{\textbf{Robustness to temperature sampling for Qwen3-1.7B.} All accuracies are percentages; bracketed values below estimates give 95\% problem-cluster bootstrap confidence intervals. Each problem has three sampled trajectories per method, each subject to the stated budget. Majority@3 requires a strict answer-level majority; Pass@3 is an oracle metric. Bold marks the higher accuracy for each metric within a benchmark.}
\label{tab:sampling-robustness}
\begin{tabular*}{\linewidth}{@{\extracolsep{\fill}}lccccc@{}}
\toprule
Benchmark & Budget & Method & \shortstack{Sampled\\pass@1} & Majority@3 & Pass@3 \\
\midrule
GSM8K & 256 & CoT-SFT & \shortstack{0.0\\{}[0.0, 0.1]} & \shortstack{0.0\\{}[0.0, 0.0]} & \shortstack{0.1\\{}[0.0, 0.2]} \\
\addlinespace[3pt]
 & 256 & GLR-10 & \shortstack{\textbf{59.9}\\{}[57.7, 62.1]} & \shortstack{\textbf{61.3}\\{}[58.6, 63.9]} & \shortstack{\textbf{74.5}\\{}[72.1, 76.7]} \\
\midrule
MATH500 & 512 & CoT-SFT & \shortstack{0.1\\{}[0.0, 0.2]} & \shortstack{0.0\\{}[0.0, 0.0]} & \shortstack{0.2\\{}[0.0, 0.6]} \\
\addlinespace[3pt]
 & 512 & GLR-10 & \shortstack{\textbf{45.0}\\{}[41.1, 48.9]} & \shortstack{\textbf{44.4}\\{}[40.0, 48.8]} & \shortstack{\textbf{57.4}\\{}[53.0, 61.8]} \\
\midrule
SVAMP & 256 & CoT-SFT & \shortstack{1.0\\{}[0.4, 1.7]} & \shortstack{0.0\\{}[0.0, 0.0]} & \shortstack{3.0\\{}[1.3, 5.0]} \\
\addlinespace[3pt]
 & 256 & GLR-10 & \shortstack{\textbf{77.8}\\{}[74.0, 81.4]} & \shortstack{\textbf{80.7}\\{}[76.0, 85.0]} & \shortstack{\textbf{91.0}\\{}[87.7, 94.0]} \\
\bottomrule
\end{tabular*}
\end{table}

\paragraph{The constrained-budget advantage persists under sampling.}
Table~\ref{tab:sampling-robustness} shows that \GLR{} retains its accuracy advantage when decoding is stochastic. The paired sampled-pass@1 differences are 59.9 percentage points [57.7, 62.1] on GSM8K, 44.9 [41.1, 48.7] on MATH500, and 76.8 [73.0, 80.4] on SVAMP. Each confidence interval lies above zero. Strict answer voting gives the same ordering: Majority@3 differences are 61.3 [58.6, 63.9], 44.4 [40.0, 48.8], and 80.7 [76.0, 85.0] percentage points, respectively, with exact paired McNemar $p<0.001$ on each benchmark. These findings show that the constrained-budget advantage persists at the evaluated sampling setting, rather than depending solely on greedy decoding. Majority@3 and Pass@3 require three trajectories per problem, so they serve as robustness measures and should not be interpreted as single-trajectory efficiency gains.

\section{Compute Accounting}
\label{app:compute-accounting}
Our primary efficiency metric counts latent steps and decoded tokens equally.
This provides a hardware-independent comparison of sequential model steps, but
it is conservative for \GLR{}. A standard decoded token requires a Transformer
forward pass followed by the full vocabulary projection and token selection.
In contrast, a latent step requires the Transformer forward pass followed only
by the transition head $g_\phi$, and does not evaluate the vocabulary head.
Thus, \GLR{} reduces generation cost through two mechanisms: it shortens the
number of sequential steps needed for successful solutions, and it reduces the
number of vocabulary-head evaluations during the latent rollout.

For Qwen3-1.7B, the vocabulary head contains roughly 300M parameters, whereas
the \GLR{} transition head contains roughly 4M parameters. Therefore, during the
first $K$ latent steps, \GLR{} replaces the large vocabulary projection with a
much smaller learned transition. This does not make a latent step equivalent to
a 4M-parameter operation, since the Transformer forward pass is still required.
However, it does mean that counting a latent step and a decoded token equally
understates the reduction in vocabulary-projection from latent rollout.


\paragraph{Measurement protocol.}
We compare Qwen3-1.7B \CoTSFT{} and \GLRH{10} on the complete MATH500 and SVAMP test sets, comprising 500 and 300 problems, respectively. Both methods use a shared Hugging Face decoder with Transformers 4.57.1, PyTorch 2.8.0, BF16, FlashAttention-2, KV caching, greedy decoding, and batch size one. Each method pair runs on the same NVIDIA H100 MIG 3g.40gb partition, with counterbalanced execution order and ten untimed warm-up examples per method on each worker. The workload is distributed over five workers, each processing requests sequentially. Generation limits are 4,096 total steps on MATH500 and 2,048 on SVAMP, counting latent and textual steps together. In this timing experiment, \GLRH{10} performs ten latent updates without evaluating the vocabulary head, followed by the shared textual decoder. We synchronize CUDA before and after generation and measure wall-clock time from prefill through the final token or stopping event. Timing excludes model loading, tokenization, detokenization, answer scoring, and file I/O. All examples contribute to the reported metrics, including incorrect and truncated generations.

\begin{table}[t]
    \centering
    \caption{\textbf{Generation latency under a shared Hugging Face decoder.} Qwen3-1.7B, batch size one, H100 MIG 3g.40gb.}
    \label{tab:matched-latency}
    \small
    \setlength{\tabcolsep}{3pt}
    \renewcommand{\arraystretch}{1.2}
    \begin{tabular*}{\linewidth}{@{\extracolsep{\fill}}lrrrrc@{}}
        \toprule
        & \multicolumn{2}{c}{Median latency (s)}
        & \multicolumn{2}{c}{P95 latency (s)}
        & Median speedup \\
        \cmidrule(lr){2-3}\cmidrule(lr){4-5}
        Benchmark & \CoTSFT{} & \GLRH{10} & \CoTSFT{} & \GLRH{10} & [95\% CI] \\
        \midrule
        MATH500 & 45.72 & \textbf{8.64} & 84.31 & \textbf{65.03} & $5.29\times$ [4.81, 5.78] \\
        SVAMP   & 10.28 & \textbf{2.24} & 40.37 & \textbf{5.71}  & $4.60\times$ [4.20, 4.90] \\
        \bottomrule
    \end{tabular*}
    \par\smallskip
    \begin{minipage}{\linewidth}
        \footnotesize Speedup is the ratio of \CoTSFT{} to \GLR{} median latency; intervals use paired bootstrap resampling.
    \end{minipage}
\end{table}

\paragraph{Shorter generations yield practical inference savings.}
Table~\ref{tab:matched-latency} shows that \GLRH{10} reduces median latency by factors of $5.29$ on MATH500 and $4.60$ on SVAMP. The corresponding confidence intervals, [4.81, 5.78] and [4.20, 4.90], both lie above one, supporting a consistent speed advantage across resampled evaluation problems. The latency tails also decrease, although less sharply on MATH500: P95 latency falls from 84.31 to 65.03 seconds, compared with 40.37 to 5.71 seconds on SVAMP. Thus, the median improvement does not imply an equal reduction for the slowest requests. Sequential throughput, computed as the number of examples divided by their summed generation times, increases from 0.0195 to 0.0774 examples/s on MATH500 and from 0.0640 to 0.3040 examples/s on SVAMP. We also release a hybrid Hugging Face and vLLM decoder that computes latent prefixes in batches and passes the prompt and latent embeddings to vLLM for textual generation. In the release-verification runs of Qwen3-1.7B \GLRH{10}, with latent-prefix batch size 64, the logged prefix and text-decoding stages take approximately $8+56$ seconds for GSM8K, $3+81$ seconds for MATH500, and $2+18$ seconds for SVAMP. The sums are approximately 64, 84, and 20 seconds for the complete benchmarks. They illustrate the practical use of the released decoder, while the controlled speedup comparison uses the shared Hugging Face stack in Table~\ref{tab:matched-latency}.

\paragraph{Compute estimates and uncertainty.}
We estimate dense-matrix-multiplication FLOPs from instrumented Transformer, vocabulary-head, and transition-head calls, counting each multiply-accumulate as two operations. Mean compute per example decreases from 9.97 to 2.58 TFLOPs on MATH500 and from 2.99 to 0.73 TFLOPs on SVAMP. These correspond to reductions of 74.1\% [71.2, 76.9] and 75.6\% [71.4, 79.0]. The estimates exclude lower-order operations and memory movement and therefore describe arithmetic work rather than total hardware cost. For latency speedups and compute reductions, we use 10,000 paired bootstrap resamples within each benchmark, retaining the \CoTSFT{}/\GLR{} pair for each sampled problem. We recompute each statistic per resample and report the 2.5th and 97.5th percentiles.

\section{Latent Trajectory Analysis Protocol}
\label{app:latent-dynamics-protocol}
We analyze Qwen3-1.7B \GLRH{10} on all 1,319 GSM8K, 500 MATH500, and 300 SVAMP examples. For each problem, the corresponding \CoTSFT{} reasoning trace is teacher-forced through the \GLR{} checkpoint, so that both the discrete reference and latent rollout are represented in the same model's final-layer hidden-state space. Endpoint metrics use the final realized latent step.

\paragraph{Contextual alignment and progress.}
Let $\mathbf{h}^{L}_{1:r}$ denote the latent hidden states and $\mathbf{h}^{C}_{1:m}$ the reference CoT hidden states for one problem. We compute a monotonic alignment by minimizing
\[
\min_{1\leq j_1\leq\cdots\leq j_r\leq m}
\sum_{k=1}^{r}\left[1-\cos(\mathbf{h}^{L}_k,\mathbf{h}^{C}_{j_k})\right].
\]
Dynamic programming finds the minimum-cost alignment, allowing repeated reference positions and leaving the starting and ending positions unconstrained. Progress at latent step $k$ is $j_k/m$. Table~\ref{tab:latent-progress-drift} reports mean final progress $100j_r/m$ as a percentage, separately for correct and incorrect generations. Since monotonicity is imposed by the alignment, increasing aligned position alone is not independent evidence of reasoning advancement; our interpretation relies on the comparison between correct and incorrect generations and the association with accuracy.

\paragraph{Embedding-space drift.}
Let $\mathbf{z}_k$ be the latent input embedding at step $k$ and $E_{\mathrm{in}}(t_{j_k})$ the input embedding of the aligned reference token. We define
\[
d^E_k=1-\cos\!\left(\mathbf{z}_k,E_{\mathrm{in}}(t_{j_k})\right).
\]
Thus, alignment is computed in contextual hidden-state space, whereas embedding drift compares input-space representations. Table~\ref{tab:latent-progress-drift} reports mean final drift $d^E_r$ within each correctness group. For the regression analysis, we also compute final hidden-state drift $d^H_r=1-\cos(\mathbf{h}^{L}_r,\mathbf{h}^{C}_{j_r})$.

\begin{figure}[t]
    \centering
    \includegraphics[width=\linewidth]{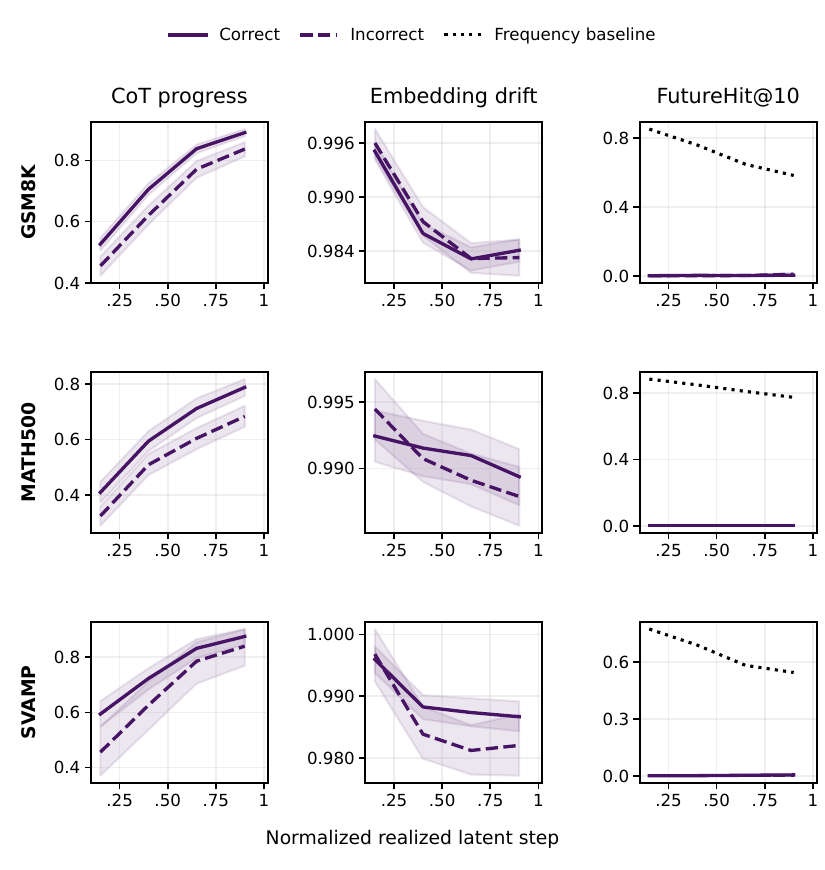}
    \caption{\textbf{Latent trajectory dynamics for Qwen3-1.7B \GLRH{10}.} Rows show GSM8K, MATH500, and SVAMP; columns show aligned CoT progress, embedding-space cosine drift, and FutureHit@10. Solid and dashed purple curves denote correct and incorrect generations, respectively. Shaded regions are pointwise 95\% percentile intervals from 2,000 bootstrap resamples of per-example bin means within each benchmark and correctness group. FutureHit@10 is plotted as a fraction rather than a percentage. Dotted black curves show the expected hit probability under ten independent draws from each benchmark's empirical reference-token frequency distribution.}
    \label{fig:latent-trajectory-curves}
\end{figure}

\paragraph{Nearest-token overlap.}
For each latent input embedding, we retrieve the ten vocabulary embeddings with highest cosine similarity, excluding tokenizer special tokens. Let $N_k$ be their token IDs and let $W_k$ contain the reference token IDs from position $j_k$ through $\min(j_k+31,m)$. The window includes the aligned token and is truncated at the end of the reference reasoning trace. The per-step hit indicator is $\mathbf{1}[N_k\cap W_k\neq\varnothing]$. FutureHit@10 is the mean of these indicators over realized latent steps within each example, followed by an average over examples. This metric measures token-identity overlap with the aligned window; it does not directly measure the semantic information encoded by a latent state. For the frequency baseline in Figure~\ref{fig:latent-trajectory-curves}, let $p(W_k)$ be the total empirical token-frequency mass of the distinct token IDs in $W_k$, estimated from all reference reasoning traces in the benchmark. The expected hit probability for ten independent draws is $1-(1-p(W_k))^{10}$. This frequency baseline uses the reference-token distribution without the special-token filtering applied to nearest-neighbor retrieval and serves as a token-frequency reference rather than a matched geometric control.

\paragraph{Evolution over the latent rollout.}
Figure~\ref{fig:latent-trajectory-curves} summarizes progress, embedding drift, and FutureHit@10 over four bins of normalized latent time $k/r$, with bin index $\lceil 4k/r\rceil$. We first average each metric over steps within each example and bin, then average across examples separately for correct and incorrect generations. Horizontal coordinates are the corresponding mean normalized latent times. Correct generations show higher mean aligned progress across the four bins on all three benchmarks, while embedding drift is similar between correctness groups and nearest-token overlap remains low. Progress is plotted as a fraction, whereas Table~\ref{tab:latent-progress-drift} reports percentages.

\paragraph{Contextual progress is associated with answer correctness.}
The differences in Table~\ref{tab:latent-progress-drift} suggest that correct generations reach later parts of the reference CoT trajectory. To examine whether this association persists after accounting for benchmark differences and other trajectory properties, we fit an unregularized logistic regression to all 2,119 examples at a latent budget of $K=10$. The model predicts generation correctness from standardized final progress, final embedding drift, final hidden-state drift, and realized latent-step count, together with categorical benchmark indicators. A one-standard-deviation increase in final progress is associated with an accuracy odds ratio of 1.31, with a 95\% confidence interval of [1.21, 1.45]. This corresponds to approximately 31\% higher odds of a correct answer, holding the other predictors fixed; it is not a 31-percentage-point increase in accuracy. The interval lies above one, supporting a positive association between later contextual alignment and answer correctness. This result complements the higher progress observed for correct generations across all three benchmarks and suggests that contextual alignment captures information relevant to successful reasoning.

\paragraph{Embedding drift provides weaker evidence of an accuracy association.}
We separately examine final embedding drift using a regression that controls for benchmark and latent-step count. A one-standard-deviation increase in drift gives an odds ratio of 1.08 [0.99, 1.19]. Unlike the progress interval, this interval includes one, so the analysis does not establish a clear association between embedding drift and correctness. It also does not establish that drift has no effect. Together with the similar drift values for correct and incorrect generations in Table~\ref{tab:latent-progress-drift}, this finding suggests that proximity to the aligned token embedding alone is insufficient to explain successful latent reasoning. The progress and drift estimates come from different regression specifications and should not be interpreted as a direct comparison of effect sizes. Both analyses describe associations within the observed trajectories; they do not show that intervening on progress or drift would change accuracy.

\paragraph{Uncertainty estimation.}
We compute 95\% percentile confidence intervals from 2,000 bootstrap fits with random seed 2026. Each resample draws benchmark/example-ID pairs with replacement from the pooled set of problems. Continuous predictors are standardized again within each bootstrap fit. We take the 2.5th and 97.5th percentiles of the fitted coefficients and exponentiate them to obtain odds-ratio intervals. These intervals quantify uncertainty from resampling evaluation problems for the fixed trained checkpoint; they do not capture variation across training runs. The shaded intervals in Figure~\ref{fig:latent-trajectory-curves} instead resample per-example bin means within each benchmark and correctness group, and are pointwise intervals rather than simultaneous confidence bands over the full trajectory.

\subsection{Alternative Latent Transitions}
\label{app:transition-controls}

\paragraph{Controlling for position advancement and transition direction.}
We compare standard \GLR{} with four inference-time controls on the same fixed subset of 100 problems from each of GSM8K, MATH500, and SVAMP, giving 300 paired examples. All conditions use the same trained Qwen3-1.7B GLR checkpoint and latent budget $K=10$; only the construction of the latent input embeddings changes. The learned condition is normal \GLR{}: the trained head predicts a displacement from the current contextual hidden state, which is added to the preceding input embedding. Each resulting input is processed at a new sequence position and retained in the KV cache. Zero transition replaces every displacement with zero, keeping the input embedding fixed while still advancing sequence positions and extending the cached context. It is therefore not equivalent to $K=0$, which skips the latent rollout entirely. Even with identical input embeddings, hidden states can change because each step occupies a different position and attends to preceding computations. Repeated \texttt{<think>} explicitly uses this opening-marker embedding at each latent position. Since our rollout starts from that same embedding, this control is equivalent to zero transition under the current initialization; their identical results are expected, rather than independent evidence from distinct mechanisms. The norm-matched random-head control replaces the learned direction with that of a randomly initialized linear head, fixed across steps and examples. Its output still depends on the current hidden state and is rescaled to the learned head's predicted update norm at that state. The noisy-transition control instead normalizes the learned update, adds a random unit direction scaled by $\sigma=1$, and renormalizes the result to restore the original update norm. It therefore retains a contribution from the learned direction while perturbing it at each step. In both direction controls, norm matching uses the current control trajectory's state, not the corresponding state of the standard GLR rollout. The stochastic controls use seed 0. Subsequent textual decoding and benchmark-specific scoring follow the same procedure across conditions.

\begin{table}[htbp]
\centering
\setlength{\tabcolsep}{4pt}
\renewcommand{\arraystretch}{1.25}
\caption{\textbf{Alternative transitions for Qwen3-1.7B at $K=10$.} Results pool 100 problems from each of three benchmarks. Median steps include all examples and latent positions. Immediate closure means the first textual token after the latent phase is \texttt{</think>}. Accuracy differences are relative to learned transitions; bracketed values give 95\% paired, benchmark-stratified bootstrap intervals from 10,000 resamples. Bold marks the highest accuracy and lowest median length.}
\label{tab:transition-controls}
\begin{tabular*}{\linewidth}{@{\extracolsep{\fill}}lcccc@{}}
\toprule
Transition & \shortstack{Accuracy\\(\%)} & \shortstack{Median\\steps} & \shortstack{Immediate\\closure (\%)} & \shortstack{$\Delta$ accuracy\\(pp)} \\
\midrule
Learned & 72.3 & \textbf{183.5} & 8.0 & \textemdash \\
\midrule
Zero & \textbf{76.0} & 948.0 & 9.7 & \shortstack{$+3.7$\\$[-1.7,\,9.0]$} \\
\midrule
Repeated \texttt{<think>} & \textbf{76.0} & 948.0 & 9.7 & \shortstack{$+3.7$\\$[-1.7,\,9.0]$} \\
\midrule
\shortstack[l]{Norm-matched\\random head} & 53.0 & 315.5 & 1.3 & \shortstack{$-19.3$\\$[-25.7,\,-13.0]$} \\
\midrule
Noisy & 58.7 & 233.0 & 13.7 & \shortstack{$-13.7$\\$[-19.3,\,-8.0]$} \\
\bottomrule
\end{tabular*}
\end{table}

\paragraph{Learned directions support the accuracy and length tradeoff.}
Table~\ref{tab:transition-controls} shows that preserving latent positions without learned movement does not reproduce the generation-length reduction: zero transitions and repeated \texttt{<think>} have a median of 948 steps, compared with 183.5 for standard \GLR{}. Their accuracy is 3.7 percentage points higher, with the paired confidence interval spanning zero. Disrupting the learned directions reduces accuracy by 19.3 percentage points for the random head and 13.7 for noisy transitions, with both intervals below zero. These results establish the role of learned directions in combining shorter generations with retained accuracy. Standard \GLR{} closes the reasoning span immediately in only 8.0\% of cases; in the remaining 92.0\%, textual reasoning continues after the latent phase. Thus, the shorter generations are not explained by a uniform immediate-closure behavior.

\section{Qualitative Examples of Latent-to-Text Transitions}
\label{app:examples}
We provide qualitative examples illustrating how \GLR{} transitions from
continuous latent updates back to explicit token generation. In each example,
\latentsteps{K} denotes the $K$ continuous latent steps performed inside the
reasoning span; these steps are not decoded into text. The tokens shown after
\latentsteps{K} are the first explicit tokens generated once standard decoding
resumes. These examples complement the quantitative results in
Section~\ref{sec:analysis}: after the latent prefix, the model often continues
from a problem-specific intermediate state rather than restarting a full
explicit chain-of-thought trace.

Figure~\ref{fig:main_examples} shows a GSM8K example from Qwen3-1.7B with
$K=20$ latent steps. The model resumes text generation with the relevant
operation, $54 - 20 = 34$, and then emits the final answer.

\begin{figure}[t]
\centering
\footnotesize
\begin{tcolorbox}[
  colback=gray!3,
  colframe=gray!45,
  boxrule=0.4pt,
  arc=2pt,
  left=3pt,right=3pt,top=3pt,bottom=3pt,
  title={GSM8K, Qwen3-1.7B, \GLRH{20}},
  fonttitle=\bfseries,
  colbacktitle=gray!25,
  coltitle=black,
  toptitle=1pt,
  bottomtitle=1pt
]
\spTok{<|im\_start|>}\textbf{user}\\
Question: The ratio of men to women on a bus is 5:9. If the total number of passengers on the bus is 84, and 20 women alight from the bus at the next stop, how many women have remained on the bus?\textbackslash nAnswer:\spTok{<|im\_end|>}\\[0.3em]
\spTok{<|im\_start|>}\textbf{assistant}\\
\spTok{<think>}\\
\latentsteps{20} 54 - 20 = 34.\\
\spTok{</think>}\\
{\ttfamily\raggedright
The ratio of men to women on the bus is 5:9. To find the number of men and women, we first determine the total parts in the ratio:  \textbackslash{}n\textbackslash{}\textbackslash{}(5 + 9 = 14\textbackslash{}\textbackslash{}).  \textbackslash{}n\textbackslash{}nSince the total number of passengers is 84, each part corresponds to:  \textbackslash{}n\textbackslash{}\textbackslash{}(\textbackslash{}\textbackslash{}frac\{84\}\{14\} = 6\textbackslash{}\textbackslash{}).  \textbackslash{}n\textbackslash{}nCalculating the number of men and women:  \textbackslash{}n- Men: \textbackslash{}\textbackslash{}(5 \textbackslash{}\textbackslash{}times 6 = 30\textbackslash{}\textbackslash{})  \textbackslash{}n- Women: \textbackslash{}\textbackslash{}(9 \textbackslash{}\textbackslash{}times 6 = 54\textbackslash{}\textbackslash{})  \textbackslash{}n\textbackslash{}nAfter 20 women alight, the remaining number of women is:  \textbackslash{}n\textbackslash{}\textbackslash{}(54 - 20 = 34\textbackslash{}\textbackslash{}).  \textbackslash{}n\textbackslash{}nThe final answer is \textbackslash{}\textbackslash{}(\textbackslash{}\textbackslash{}boxed\{34\}\textbackslash{}\textbackslash{})
\par}
\spTok{<|im\_end|>}
\end{tcolorbox}
\caption{
\textbf{Qualitative example of latent-to-text transition.}
A Qwen3-1.7B \GLRH{20} generation on GSM8K. After 20 latent steps inside the reasoning span, the model resumes explicit text generation
mid-solution, directly using the relevant quantity and operation
($54 - 20 = 34$) before emitting the final answer. We display the continuous
updates as \latentsteps{20}; special tokens are shown in teal and latent-step
placeholders in orange.
}
\label{fig:main_examples}
\end{figure}

\begin{figure}[t]
\centering
\footnotesize
\begin{tcolorbox}[
  colback=gray!3,
  colframe=gray!45,
  boxrule=0.4pt,
  arc=2pt,
  left=3pt,right=3pt,top=3pt,bottom=3pt,
  title={GSM8K, Qwen3-1.7B, \GLRH{50}},
  fonttitle=\bfseries,
  colbacktitle=gray!25,
  coltitle=black,
  toptitle=1pt,
  bottomtitle=1pt
]
\spTok{<|im\_start|>}\textbf{user}\\
Question: The ratio of men to women on a bus is 5:9. If the total number of passengers on the bus is 84, and 20 women alight from the bus at the next stop, how many women have remained on the bus?\textbackslash nAnswer:\spTok{<|im\_end|>}\\[0.3em]
\spTok{<|im\_start|>}\textbf{assistant}\\
\spTok{<think>}\\
\latentsteps{50} \\
\spTok{</think>}\\
{\ttfamily\raggedright
he ratio of men to women on the bus is 5:9. The total number of passengers is 84. First, we calculate the number of men and women:\textbackslash{}n\textbackslash{}n1. The total parts in the ratio are \textbackslash{}\textbackslash{}(5 + 9 = 14\textbackslash{}\textbackslash{}).\textbackslash{}n2. Number of men: \textbackslash{}\textbackslash{}(\textbackslash{}\textbackslash{}frac\{5\}\{14\} \textbackslash{}\textbackslash{}times 84 = 30\textbackslash{}\textbackslash{}).\textbackslash{}n3. Number of women: \textbackslash{}\textbackslash{}(\textbackslash{}\textbackslash{}frac\{9\}\{14\} \textbackslash{}\textbackslash{}times 84 = 54\textbackslash{}\textbackslash{}).\textbackslash{}n\textbackslash{}nAt the next stop, 20 women alight. The remaining women on the bus are:\textbackslash{}n\textbackslash{}\textbackslash{}[54 - 20 = 34\textbackslash{}\textbackslash{}].\textbackslash{}n\textbackslash{}n\textbackslash{}\textbackslash{}boxed\{34\}
\par}
\spTok{<|im\_end|>}
\end{tcolorbox}
\caption{
\textbf{Qualitative example of latent-to-text transition.}
A Qwen3-1.7B \GLRH{50} generation on GSM8K. We display the continuous
updates as \latentsteps{50}; special tokens are shown in teal and latent-step
placeholders in orange.
}
\label{fig:main_examples_3}
\end{figure}

\begin{figure}[t]
\centering
\footnotesize
\begin{tcolorbox}[
  colback=gray!3,
  colframe=gray!45,
  boxrule=0.4pt,
  arc=2pt,
  left=3pt,right=3pt,top=3pt,bottom=3pt,
  title={GSM8K, Qwen3-1.7B, \GLRH{80}},
  fonttitle=\bfseries,
  colbacktitle=gray!25,
  coltitle=black,
  toptitle=1pt,
  bottomtitle=1pt
]
\spTok{<|im\_start|>}\textbf{user}\\
Question: The ratio of men to women on a bus is 5:9. If the total number of passengers on the bus is 84, and 20 women alight from the bus at the next stop, how many women have remained on the bus?\textbackslash nAnswer:\spTok{<|im\_end|>}\\[0.3em]
\spTok{<|im\_start|>}\textbf{assistant}\\
\spTok{<think>}\\
\latentsteps{80} \\
\spTok{</think>}\\
{\ttfamily\raggedright
The ratio of men to women on the bus is 5:9. The total number of passengers is 84. First, we calculate the total parts in the ratio:\textbackslash{}n\textbackslash{}n\textbackslash{}\textbackslash{}[\textbackslash{}n5 + 9 = 14 \textbackslash{}\textbackslash{}text\{ parts\}\textbackslash{}n\textbackslash{}\textbackslash{}]\textbackslash{}n\textbackslash{}nEach part corresponds to:\textbackslash{}n\textbackslash{}n\textbackslash{}\textbackslash{}[\textbackslash{}n84 \textbackslash{}\textbackslash{}div 14 = 6\textbackslash{}n\textbackslash{}\textbackslash{}]\textbackslash{}n\textbackslash{}nThus, the number of men is \textbackslash{}\textbackslash{}(5 \textbackslash{}\textbackslash{}times 6 = 30\textbackslash{}\textbackslash{}) and the number of women is \textbackslash{}\textbackslash{}(9 \textbackslash{}\textbackslash{}times 6 = 54\textbackslash{}\textbackslash{}).\textbackslash{}n\textbackslash{}nAt the next stop, 20 women alight. The remaining number of women is:\textbackslash{}n\textbackslash{}n\textbackslash{}\textbackslash{}[\textbackslash{}n54 - 20 = 34\textbackslash{}n\textbackslash{}\textbackslash{}]\textbackslash{}n\textbackslash{}n\textbackslash{}\textbackslash{}[\textbackslash{}n\textbackslash{}\textbackslash{}boxed\{34\}\textbackslash{}n\textbackslash{}\textbackslash{}]
\par}
\spTok{<|im\_end|>}
\end{tcolorbox}
\caption{
\textbf{Qualitative example of latent-to-text transition.}
A Qwen3-1.7B \GLRH{80} generation on GSM8K. We display the continuous
updates as \latentsteps{80}; special tokens are shown in teal and latent-step
placeholders in orange.
}
\label{fig:main_examples_2}
\end{figure}

\section{Future Work}
\label{app:future}
Future work should extend \GLR{} beyond 8B parameters and use substantially larger reasoning datasets; our experiments use training subsets of 5.4K or 10K examples.
Another promising direction is to replace deterministic path approximation with diffusion- or flow-style latent trajectory modeling, where test-time scaling can be viewed as sampling multiple points or paths in continuous reasoning space. Finally, \GLR{} should be evaluated beyond mathematics, including science, code generation, theorem proving, planning, and multi-hop reasoning, to test whether the observed compression effect is domain-general.

\section{Limitations}
\label{app:limitation}
This study is limited by training scale: we evaluate Qwen3 models up to 8B parameters using only 5.4K or 10K CoT examples, which may understate the potential of \GLR{} and contribute to drift at large latent budgets.
Our evaluation is also focused on mathematical reasoning, so it remains unclear whether the same accuracy--length tradeoff holds for other domains such as science, code, and general reasoning. Finally, replacing explicit reasoning with continuous states makes part of the model's computation less interpretable.
Our controlled latency measurements cover Qwen3-1.7B at batch size one on MATH500 and SVAMP using H100 MIG partitions; performance across other hardware, model sizes, and serving workloads remains to be evaluated systematically.

\section{Broader Impacts}
\label{app:broader}
This work frames latent reasoning as a geometric path-approximation problem in a model's pretrained token-embedding space, and shows that \GLR{} can replace early explicit CoT tokens with continuous latent steps to induce shorter generations. This perspective may help reduce the inference cost of reasoning-heavy LLMs by exposing an explicit tradeoff between latent computation budget, output length, and accuracy. However, because \GLR{} hides part of the reasoning trajectory inside non-verbalized embedding-space states, it may make model behavior harder to inspect than standard CoT decoding. Future uses of \GLR{}-style methods should therefore pair token-efficient latent inference with careful answer verification, drift monitoring, and tools for interpreting latent-to-text transitions.

\newpage
\section*{NeurIPS Paper Checklist}

\begin{enumerate}

\item {\bf Claims}
    \item[] Question: Do the main claims made in the abstract and introduction accurately reflect the paper's contributions and scope?
    \item[] Answer: \answerYes{} 
    \item[] Justification: We discuss experimental results to validate our proposed idea.

\item {\bf Limitations}
    \item[] Question: Does the paper discuss the limitations of the work performed by the authors?
    \item[] Answer: \answerYes{} 
    \item[] Justification: The authors provide sufficient information regarding limitations fo the current work in a separate section.

\item {\bf Theory assumptions and proofs}
    \item[] Question: For each theoretical result, does the paper provide the full set of assumptions and a complete (and correct) proof?
    \item[] Answer: \answerYes{} 
    \item[] Justification: The authors provide for all theoretical results the full set of assumptions
accompanied by their proofs.

    \item {\bf Experimental result reproducibility}
    \item[] Question: Does the paper fully disclose all the information needed to reproduce the main experimental results of the paper to the extent that it affects the main claims and/or conclusions of the paper (regardless of whether the code and data are provided or not)?
    \item[] Answer: \answerYes{} 
    \item[] Justification: The authors provide all the necessary information need to reproduce the experiments, including hyperparameters, optimizer, model architectures, etc. in the appendix.

\item {\bf Open access to data and code}
    \item[] Question: Does the paper provide open access to the data and code, with sufficient instructions to faithfully reproduce the main experimental results, as described in supplemental material?
    \item[] Answer: \answerYes 
    \item[] Justification: The code is publicly released at: \url{https://github.com/idiap/GLR}

\item {\bf Experimental setting/details}
    \item[] Question: Does the paper specify all the training and test details (e.g., data splits, hyperparameters, how they were chosen, type of optimizer) necessary to understand the results?
    \item[] Answer: \answerYes{} 
    \item[] Justification: The training and test details are provided throughout the paper and any
remaining details and hyperparameters are provided in the appendix.

\item {\bf Experiment statistical significance}
    \item[] Question: Does the paper report error bars suitably and correctly defined or other appropriate information about the statistical significance of the experiments?
    \item[] Answer: \answerYes 
    \item[] Justification: We report statistical significance analyses for selected experiments in the appendix. The remaining experiments use a fixed seed due to computational constraints.

\item {\bf Experiments compute resources}
    \item[] Question: For each experiment, does the paper provide sufficient information on the computer resources (type of compute workers, memory, time of execution) needed to reproduce the experiments?
    \item[] Answer: \answerYes{} 
    \item[] Justification: Compute details are provided in appendix.
    
\item {\bf Code of ethics}
    \item[] Question: Does the research conducted in the paper conform, in every respect, with the NeurIPS Code of Ethics \url{https://neurips.cc/public/EthicsGuidelines}?
    \item[] Answer: \answerYes{} 
    \item[] Justification: The paper raises no ethical concerns and complies with the NeurIPS guidelines.
No high-risk applications or data are involved, and fairness and privacy considerations are
acknowledged where applicable.

\item {\bf Broader impacts}
    \item[] Question: Does the paper discuss both potential positive societal impacts and negative societal impacts of the work performed?
    \item[] Answer: \answerYes{} 
    \item[] Justification: The authors include a Broader Impacts section in the appendix.
    
\item {\bf Safeguards}
    \item[] Question: Does the paper describe safeguards that have been put in place for responsible release of data or models that have a high risk for misuse (e.g., pre-trained language models, image generators, or scraped datasets)?
    \item[] Answer: \answerNA{} 
    \item[] Justification: The models released do not pose potential for misuse.

\item {\bf Licenses for existing assets}
    \item[] Question: Are the creators or original owners of assets (e.g., code, data, models), used in the paper, properly credited and are the license and terms of use explicitly mentioned and properly respected?
    \item[] Answer: \answerYes{} 
    \item[] Justification: All datasets and code used are properly cited, and licenses are referenced.
There is no evidence of license violations.

\item {\bf New assets}
    \item[] Question: Are new assets introduced in the paper well documented and is the documentation provided alongside the assets?
    \item[] Answer: \answerYes{} 
    \item[] Justification: Any new assets introduced include usage instructions, and documentation
appears sufficient for independent use.

\item {\bf Crowdsourcing and research with human subjects}
    \item[] Question: For crowdsourcing experiments and research with human subjects, does the paper include the full text of instructions given to participants and screenshots, if applicable, as well as details about compensation (if any)? 
    \item[] Answer: \answerNA{} 
    \item[] Justification: There are no experiments related to crowdsourcing and research with human
subjects.

\item {\bf Institutional review board (IRB) approvals or equivalent for research with human subjects}
    \item[] Question: Does the paper describe potential risks incurred by study participants, whether such risks were disclosed to the subjects, and whether Institutional Review Board (IRB) approvals (or an equivalent approval/review based on the requirements of your country or institution) were obtained?
    \item[] Answer: \answerNA{} 
    \item[] Justification: There is no research with human subjects.

\item {\bf Declaration of LLM usage}
    \item[] Question: Does the paper describe the usage of LLMs if it is an important, original, or non-standard component of the core methods in this research? Note that if the LLM is used only for writing, editing, or formatting purposes and does \emph{not} impact the core methodology, scientific rigor, or originality of the research, declaration is not required.
    \item[] Answer: \answerNA{} 
    \item[] Justification: \answerNA{}

\end{enumerate}

\end{document}